\documentclass[nohyperref]{article}

\usepackage{booktabs} 
\usepackage{graphicx}
\usepackage{microtype}
\usepackage{multicol} 
\usepackage{multirow} 
\usepackage{subfigure}

\usepackage{amsmath}
\DeclareMathOperator*{\ave}{avg}

\usepackage{hyperref}

\usepackage[accepted]{icml2022}

\usepackage{amsmath}
\usepackage{amssymb}
\usepackage{mathtools}
\usepackage{amsthm}
 \usepackage{nicematrix}
\usepackage{enumitem}
\usepackage[capitalize,noabbrev]{cleveref}

\usepackage[textsize=tiny]{todonotes}
\usepackage{xcolor}         

\usepackage{pdflscape}
\usepackage{afterpage}

\theoremstyle{plain}

\theoremstyle{definition}

\theoremstyle{remark}

\icmltitlerunning{On Robustness in Multimodal Learning}

\begin{document}

\twocolumn[
\icmltitle{On Robustness in Multimodal Learning}



\icmlsetsymbol{equal}{*}

\begin{icmlauthorlist}
\icmlauthor{Brandon McKinzie}{yyy}
\icmlauthor{Joseph Cheng}{zzz}
\icmlauthor{Vaishaal Shankar}{yyy}
\icmlauthor{Yinfei Yang}{xxx}
\icmlauthor{Jonathon Shlens}{yyy}
\icmlauthor{Alexander Toshev}{yyy}
\end{icmlauthorlist}

\icmlaffiliation{xxx}{Apple}
\icmlaffiliation{yyy}{Apple ML Research}
\icmlaffiliation{zzz}{Work done while at Apple}

\icmlcorrespondingauthor{Alexander Toshev}{toshev@apple.com}

\icmlkeywords{Machine Learning, ICML}


\vskip 0.3in
]



\printAffiliations  

\begin{abstract}

Multimodal learning is defined as learning over multiple heterogeneous input modalities such as video, audio, and text. In this work, we are concerned with understanding how models behave as the type of modalities differ between training and deployment, a situation that naturally arises in many applications of multimodal learning to hardware platforms. We present a multimodal robustness framework to provide a systematic analysis of common multimodal representation learning methods. Further, we identify robustness short-comings of these approaches and propose two intervention techniques leading to $1.5\times$-$4\times$ robustness improvements on three datasets, AudioSet, Kinetics-400 and ImageNet-Captions. Finally, we demonstrate that these interventions better utilize additional modalities, if present, to achieve competitive results of $44.2$ mAP on AudioSet 20K.
\end{abstract}
\section{Introduction}
\label{sec:introduction}

Machine learning models in the real world operate on a wide range of hardware platforms and sensor suites. Deployed models must operate on platforms ranging from wearable devices to autonomous vehicles in which a diverse suite of sensors provide a continuous commentary about the environment.
Building a traditional machine learning model in this setting is challenging because {\it jointly} measuring data across {\it all} sensors might be infeasible. Likewise, sensor modalities may be added (or fail) at any time indicating that the tacit assumption of {\it i.i.d.} data may not occur in the real world.

Hence, properties of robustness across modalities become paramount when deploying a machine learning system to operate in a multimodal setting. First, a model should be able to operate on modalities not explicitly observed during training. For instance, we hope that the presence of additional modalities with no explicit labels may still benefit overall predictive performance. Second, models should gracefully degrade in the absence of modalities at test time. Both properties are unique to the multimodal setting.

\begin{figure}[t]
    \centering
    \includegraphics[width=0.45\textwidth]{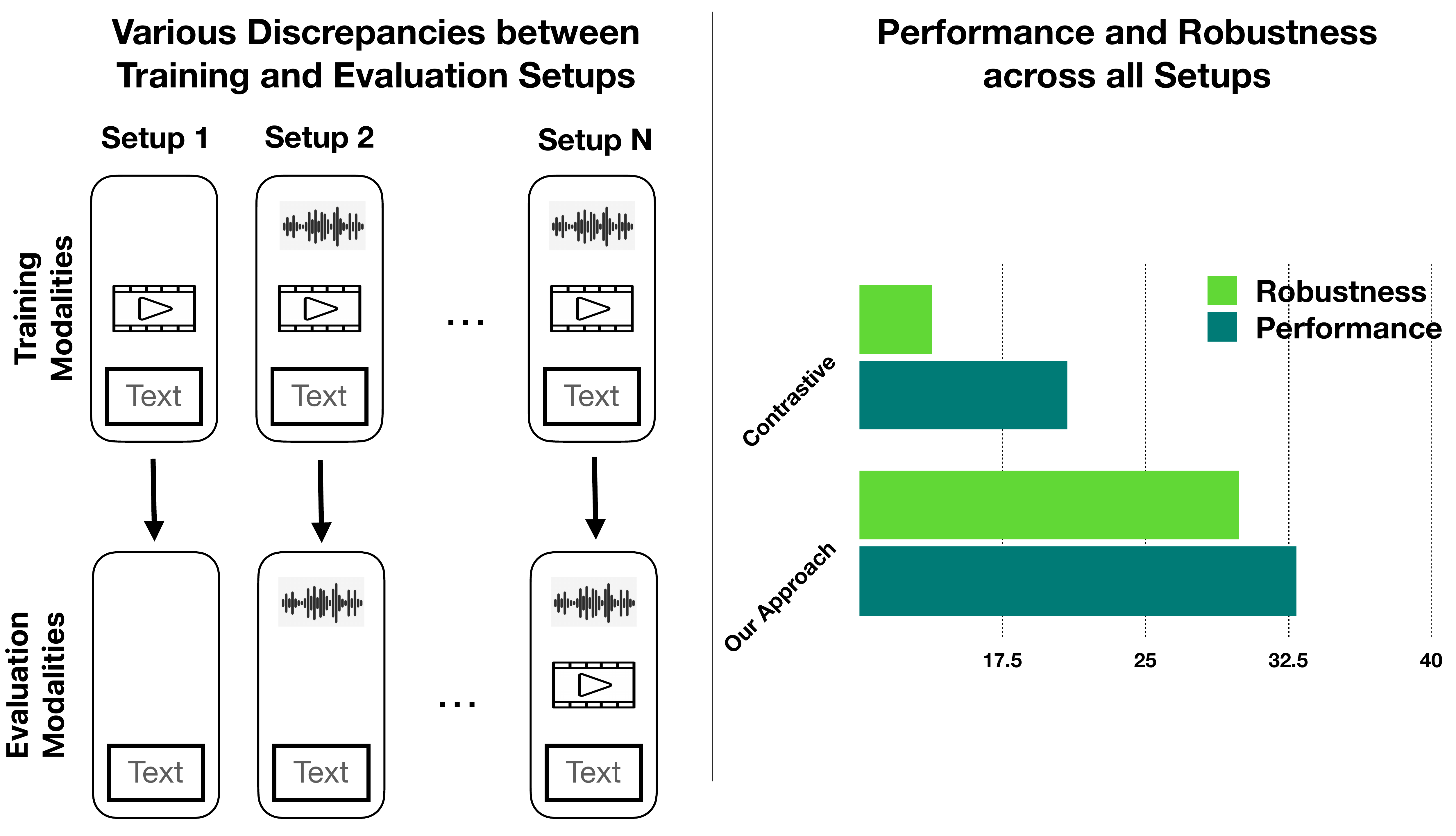}
    \vspace{-0.2cm}
    \caption{{\bf Multimodal experimental setup and results.} We study representation learning for multimodal models which exhibit discrepancies between training and evaluation modalities. We define an analysis framework for this setup, study existing pretraining methods, and propose methods to improve robustness and performance.}
    \label{fig:intro_figure}
    \vspace{-0.4cm}
\end{figure}

To address these challenges, we study the problem of multimodal robustness. {\it How do models behave when arbitrary combinations of modalities may be added or removed at test time?} 
Supervised learning typically trains a model on a labeled dataset and examines how performance deteriorates as the hold-out validation set diverges from the training set \cite{recht2019imagenet,Shankar_2021_ICCV,hendrycksCorruption}. In our setting, we wish to instead build models in which one may flexibly swap in or out individual modalities that the model has seen during pretraining, downstream training, or both (Fig. \ref{fig:intro_figure}).

One approach for achieving a flexible and performant representation
to a suite of modalities is to have a model to learn a shared representation invariant to the modality identities -- and subsequently train a discriminative model on top of that learned representation \cite{goodfellow2016deep}. Several approaches for learning a shared representation have been explored in the literature, but recently, two prominent approaches -- masked autoencoders \cite{gong2022contrastive, geng2022multimodal} and contrastive learning \cite{radford2021clip, wu2022large} -- have demonstrated extraordinary promise in the setting of multimodal representations \cite{akbari2021vatt}. We focus our work on benchmarking robustness in representation learning, and ask how to improve such representations through new training strategies.

In this work we introduce a framework for measuring robustness in multimodal settings. We define a new robustness metric to capture variability across modalities by focusing on both average and worst-case performance across training and evaluation setups. Furthermore, we stratify these metrics across common scenarios such as adding, dropping, or completely swapping the modalities fed into the model.

We focus our experiments on representation learning with the AudioSet dataset \cite{gemmeke2017audio} in which three prominent modalities -- audio, video and text -- may be systematically manipulated. Additionally, we explore the generality of our results on Kinetics-400 \cite{kay2017kinetics} and ImageNet-Captions \cite{imagenet-captions}.



We measure average and worst case performance when modalities are added or dropped at test time. To alleviate these degradations, we introduce two approaches to improve representation learning in a multimodal setting. The first approach --- derived from knowledge distillation \cite{hinton2015distilling} -- termed \textit{Modality Augmented Self-Distillation} (MASD), encourages consistency in the learned representations across labeled and unlabeled modalities. 
The second approach, derived from WiseFT~\cite{Wortsman_2022_CVPR}, leverages a weighted combination of finetuned downstream weights and the initialization pretrained weights to induce robustness.
We summarize our contributions as follows:
\setlist{nolistsep}
\begin{enumerate}[noitemsep,leftmargin=*]
\item Introduce metrics and characterize performance in a multimodal setting on several  datasets in terms of worst and average case performance.
\item Demonstrate training interventions (e.g.~MASD, WiseFT) may additively lead to $1.5\times$-$4\times$ improvement of robustness on AudioSet, Kinetics-400 and ImageNet-Captions.
\item Increasing the number of modalities used to learn a representation improves downstream performance. In particular, we obtain SOTA results ($44.2$ mAP) on AudioSet-20K by leveraging text as an additional pretraining modality.
\end{enumerate}
We hope that these results may accelerate the field of multimodal learning by offering simple, standard metrics and strong benchmarks for future improvements.
\section{Related Work}
\subsection{Robustness}
Robust machine learning has been a subject of study for decades. The support vector machine algorithm was presented as a ``robust" prediction method \cite{boser1992training} by finding the maximum margin classifier. Recently however there has been a push towards more practical forms of robustness for models operating on vision, natural language, speech and other modalities.  

Worst case adversarial examples have been extensively studied in many domains \cite{szegedy2013intriguing, alzantot2018generating, carlini2018audio} and while many effective ``defense" methods have been proposed \cite{madry2017towards, carliniAdvCert, carmonDefense} it has been shown that these defenses reduce benign (non-adversarial) accuracy and don't generalize to \emph{other} more natural forms of robustness \cite{taori2020}. A similar story arises with synthetic ``corruption" robustness \cite{hendrycksCorruption, Geirhos2018} where robust methods have been proposed but they fail to generalize to non synthetic corruptions.

For the class of \textit{natural} corruptions or distribution shifts recent large scale multimodal image-text models~\cite{radford2021clip, Pham2021} have shown unprecedented robustness when evaluated in a \textit{zero-shot} manner~\cite{recht2019imagenet, barbu2019objectnet, Shankar_2021_ICCV, gu2019}. Subsequent work has demonstrated improvements for robustness in fine-tuned models \cite{Wortsman_2022_CVPR}.

\subsection{Multimodal Learning}
A natural way to learn a representation in a self-supervised manner from streams of multimodal data is to (1) have a set of modality encoders and an aggregator producing a single representation from all available modalities and (2) to consider paired modalities as positive examples. This way of thinking naturally lends itself to contrastive learning that embeds different modalities in a common space \citep{radford2021clip, sohn2016npair}. Most of the current work focuses on image and text only \cite{radford2021clip,alayrac2022flamingo,yuan2021florence,you2022msclip} with a number of recent efforts in including video, audio, and even tabular data \citep{akbari2021vatt, alayrac2020versatile, liang2022highmmt}.

An alternative to contrastive learning is a masked reconstruction objective. Most previous approaches have focused on single modalities, such as text \cite{devlin2018bert}, images \cite{he2022masked}, videos \cite{feichtenhofer2022masked}, and audio \citep{baade2022mae, chong2022maskspec}. More recently, this approach has also been adopted in multimodal settings \cite{geng2022multimodal, wang2022image}. Other works employ both masked reconstruction and contrastive objectives \cite{gong2022contrastive,yang2022code,fang2022eva,singh2021flava}.

From a model architecture perspective, it remains an open question how best to fuse information from different modalities \citep{dou2021empiricalvisionandlanguage,liu2018learntocombine}. The flexibility of transformers \cite{vaswani2017attention} enables them to be readily adapted to other modalities beyond language \cite{akbari2021vatt,liang2022highmmt,jaegle2021perceiver,nagrani2021attention,yang2022code}.

Increasing the number of modalities poses a challenge in training and in understanding the models. In supervised learning, the greedy nature of learning can be observed and quantified \citep{wu2022characterizing,hessel2020does}, as well as intra-modality and inter-modality heterogeneity \cite{liang2022highmmt}. 











\section{Evaluation of Multimodal Representations}\label{sec:robustness_framework}
\begin{figure}
    \centering
    \includegraphics[width=0.9\columnwidth]{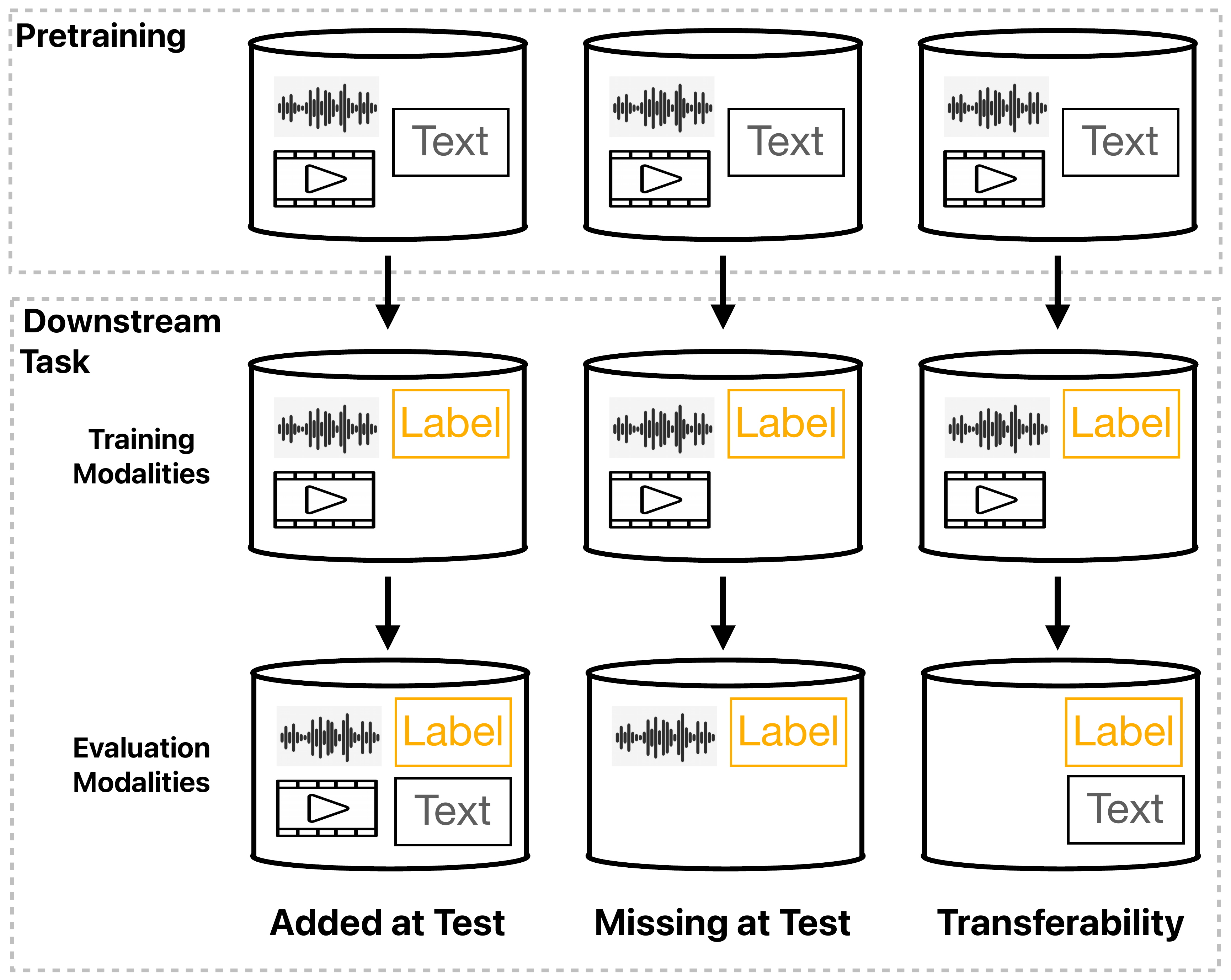}
    \caption{{\bf Multimodal experiment setup:} pretraining, downstream task training, and evaluation (see Sec.~\ref{sec:robustness_framework}), using as an example three modalities: video, audio, and text. The task at hand is classification, hence the presence of a label. At pretraining all modalities are present, while for a task only a subset is present. We describe three important setup corresponding to whether the evaluation contains more or less modalities, or a completely different set of modalities than at training: \textit{Missing at Test}, \textit{Added at Test}, and \textit{Transferability} (see Sec.~\ref{sec:analysis}).}
    \label{fig:multimodal_setup}
\end{figure}
\subsection{Setup and Notation}\label{sec:notation}
In this work we make several assumptions for our data that hold for a wide range of applications (see Fig.~\ref{fig:multimodal_setup}). First, we assume that we have readily available multimodal data consisting of several parallel input streams of different aligned modalities. Second, the above data can be acquired independently of the tasks of interest, although it might be related to it, and thus does not contain supervision. 

We will refer to these data as \textit{unsupervised pretraining data} $D$ and the set of $n$ modalities present in it by $M=\{m_1,\ldots, m_n\}$. Since we focus on subsets of modalities, it will be useful to refer to the data points $x$ and datasets $D$ restricted to a set of modalities $m\subseteq M$ by:
\begin{equation}
x |_m \quad \textrm{and} \quad D|_m = \{x|_m, x \in D\}
\end{equation}

Further, for a downstream task we have data with supervision for both training and evaluation. It is reasonable to expect that the data with supervision are substantially smaller in quantity than the pretraining data. We refer to these data as \textit{downstream training data} $D_T$ with training modalities $M_T\subseteq M$, and \textit{downstream evaluation data} $D_E$ with evaluation modalities $M_E\subseteq M$. Importantly, the training and evaluation modality sets are allowed to be different, $M_T\neq M_E$, leading to robustness issues as shown later.

{\bf Downstream Task}. Denote by $f_\theta(x)$ the downstream task model with weights $\theta$. Note that $f$ is multimodal, i.e. it can be applied on any subset $m\subseteq M$ of modalities, and such an application is denoted by $f_\theta(x|_m)$. 

The parameters of the model are estimated by training for the downstream task on $D_T$ using a task specific loss $L$:
\begin{equation}\label{eq:task_loss}
L_\textrm{task}(D_T|_{M_T})=\sum_{x\in D_T} L(f_\theta(x|_{M_T})
\end{equation}
where we explicitly say that the model is applied on $x$ using only the modalities in $M_T$.

\subsection{Multimodal Robustness Metrics}
\label{subsec:metrics}

 It is fair to assume that the downstream task of interest has a well established performance score $p$ that can be measured for our model $f$. If this score is computed on the evaluation data $D_E|_{M_E}$ using modalities $M_E$ after the model has been trained on $D_T|_{M_T}$ using modalities $M_T$, we denote this performance score by $p(M_E; M_T)$, where for brevity we skip the model and dataset notation. 

Given a set of training modalities $M_T$, we propose to measure two aspects across all evaluation setups. The first is the average score, called performance, and represents how well the modalities $M_T$ train a model when evaluated across all possible circumstances: 
\begin{align}\label{eq:performance_per_training_mods}
    P(M_T) = \ave_{M_E \subseteq M} p(M_E; M_T)
\end{align}
The second is the  the worst score, called robustness, representing the worst possible deployment scenario for the model trained on $M_T$:
\begin{align}\label{eq:robustness_per_training_mods}
    R(M_T) = \min_{M_E \subseteq M} p(M_E; M_T)
\end{align}
To produce a single set of metrics for a model across all possible training setups $M_T$, we propose to aggregate the above average and worst case performances in two ways. First, if one has control over picking an optimal training set, it makes sense to find the best performance and robustness. If we would like to evaluate on all possible training sets, then it makes sense to compute the average across training setups. We will refer to the former metrics as best Performance ($P_\textrm{best}$) and Robustness ($R_\textrm{best}$), and to the latter as simply Performance ($P$) and Robustness ($R$):
\begin{align}
     & P_\textrm{best} = \max_{M_T \subseteq M}P(M_T), \quad
    R_\textrm{best} = \max_{M_T \subseteq M}R(M_T)  \label{eq:best_metrics}\\
     & P = \ave_{M_T \subseteq M}P(M_T), \quad
    R = \ave_{M_T \subseteq M}R(M_T)   \label{eq:metrics}
\end{align}

{\bf Stratification of Performance and Robustness}. The above metrics are originally defined over all possible evaluation modality sets $M_E\subseteq M$ for each training set . However, as motivated in Sec.~\ref{sec:introduction} there can be various types of discrepancies. To better capture this, we refine $P(M_T)$ and $R(M_T)$ to be computed over a subset of possible evaluation modality sets $M_E$ (Fig.~\ref{fig:multimodal_setup}):
\setlist{nolistsep}
    \begin{enumerate}[noitemsep,leftmargin=*]
    \item \textbf{Missing at Test}: Testing modalities are a strict subset of the training modalities: $M_E \subset M_T$. This setup corresponds to having incomplete information at test time.
    
    \item \textbf{Added at Test}: Testing modalities are a strict superset of the training modalities: $M_T \subset M_E$. This setup corresponds to modalities not present during training.
    
    \item \textbf{Transferability}: Testing and training modalities are completely distinct: $M_T\cap M_E= \emptyset $. This is the most extreme setup, and tests the ability to transfer a task learned on one set to a completely different set of modalities.
\end{enumerate}
We impose the above constraints on $M_T$ and $M_E$ in the computation of $P$ and $R$ in Eq.~(\ref{eq:performance_per_training_mods}) and Eq.~(\ref{eq:robustness_per_training_mods}), and by proxy in Eq.~(\ref{eq:robustness_per_training_mods}). 

Note that when the data has only two modalities, i.e. $|M|=2$, for \textit{Added at Test} and \textit{Transferability} robustness and performance are identical $R=P$, as for every training modality set, there is only one evaluation modality set satisfying \textit{Added at Test} and \textit{Transferability} combinations. Then, the average and minimum opertions in Eq.~(\ref{eq:performance_per_training_mods}) and Eq.~(\ref{eq:robustness_per_training_mods}) result in the same values. 
\section{Multimodal Self-Supervised Learning}\label{sec:analysis}
\subsection{Models}\label{sec:standard_models}
{\bf Pretraining}. Multimodal data, as paired streams of different modalities, is a natural candidate for self-supervised learning as it is reasonable to assume that different modalities present different views of the same underlying content. This can be operationalized using contrastive~\cite{radford2021clip, jia2021scaling} or masked reconstruction~\cite{he2022masked} objectives.

For the multimodal setup, we encode the different modalities with modality-specific encoders. In the case of contrastive learning, we follow closely the VATT architecture by \citet{akbari2021vatt}, and formulate pair-wise InfoNCE losses~\cite{gutmann2010noise,oord2018representation} across all possible pairs of input modalities. This objective tries to learn per-modality representations that are as similar as possible for paired modalities. For MAE, we closely follow the AV-MAE baseline architecture described in \citet{gong2022contrastive}. Although masked reconstruction does not explicitly enforce a shared representation space for modalities, the hope is that the final shared-modality encoder layer contains information transferable from one modality to another. For further details of the formulation as well as architecture, we refer the reader to the Appendix and Sec.~\ref{sec:experiments}.

{\bf Downstream Training}. After learning a representation using SSL, we apply it for a downstream task. In particular, denote by $E_i$ the encoder for modality $m_i$ that embeds an input $x|_{m_i}$ of this modality into a Euclidean space $E_i(x|_{m_i})\in\mathbb{R}^d$ (see Sec.~\ref{sec:notation} for notation). Suppose, at downstream training or inference time, the data $D|_{M'}$ have a subset of modalities $M'\subseteq M$. Then, the final representation for $x\in D$:
\begin{equation}
    E(x) = \frac{1}{|M'|}\sum_{m'\in M'} E_{m'}(x|_{m'})
\end{equation}
This representation is used, for example in the case of a classification downstream task, to learn a classifier.


\subsection{Improving Multimodal Robustness}\label{sec:interventions}
 We hypothesize that during downstream task learning, we see only a subset of all possible modalities, and as such this learning can `damage' the pretrained model and diminish its ability to deal with the modalities not seen during downstream training. To address this challenge, we propose to apply ideas from transfer learning.

\subsubsection{Modality Augmented Self-Distillation}
\begin{figure}[t]
    \centering
    \includegraphics[width=0.4\textwidth]{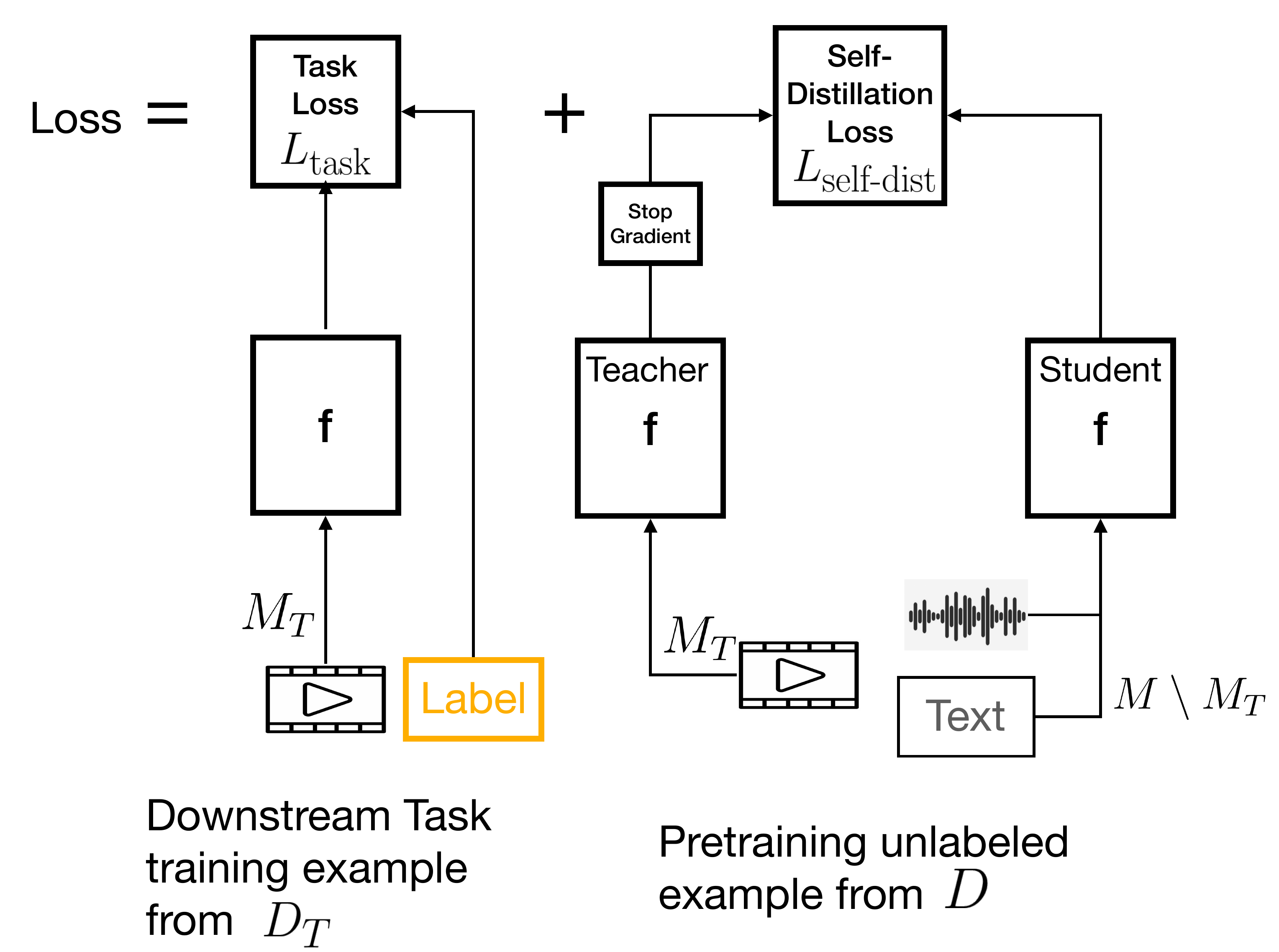}
    \caption{Diagram of Modality Augmented Self-Distillation. The Downstream task loss on the left receives labeled examples with $M_T$ modalities (in this example Video), while the self-distillation loss receives unlabeled examples with all modalities, $M_T$ are routed to the teacher network and $M\setminus M_T$ to the student (in this example, Audio and Text).}
    \label{fig:masd}
    \vspace{-0.4cm}
\end{figure}
One way to mitigate the problem is to use the pretraining data that contain all modalities but no supervision. These data can be used to regularize the performance of the model on all the modalities, even if this model is trained with a subset of the modalities present in the downstream training data. To achieve this, we draw inspiration from \cite{li2017learning,castro2018end,hou2018lifelong,rebuffi2017icarl} to use Knowledge Distillation~\cite{hinton2015distilling} on the pretraining data. 

In more detail, assume that the downstream task is classification and the model $f_\theta(y, x)$ produces probabilities over labels $y$ for a given input $x$. Then, the teacher model $f_\theta(y,x|_{M_T})$ is the same model trained over the downstream training modalities $M_T$ and data $D_T$. The student model is the same model $f_\theta(y,x|_{M \setminus M_T})$ as well (same weights), however, restricted over the modalities $M \setminus M_T$ \textit{not} present in the downstream training data. Since the student and teacher models share the same weights, but have different input modalities, we call this loss self-distillation:
\begin{equation*}
    L_{\textrm{self-dist}}(D) = -\sum_{x\in D} \sum_y f_\theta(y, x|_{M_T}) \log\left(f_\theta(y, x|_{M \setminus M_T})\right)
\end{equation*}

The final objective of MASD combines the above loss with the downstream task loss from Eq.~(\ref{eq:task_loss}) (see Fig.~\ref{fig:masd}):
\begin{equation}\label{eq:masd_loss}
L_{MASD} = L_\textrm{task}(D_T|_{M_T}) + L_{\textrm{self-dist}}(D_{SD})
\end{equation}
where the self-distillation loss is defined of a subset $D_{SD}\subset D$ of the pre-training data.

Since both the student and teacher model share the same weights $\theta$, the above loss makes sure that the model is well behaved across all modalities $M$. Note that for training stability we stop the gradient flow through the teacher.

\subsubsection{Applying WISE-FT to MASD models}
There has been a recent line of work on improving the distributional robustness of finetuned large scale image-text models by \emph{weight-space ensembling} (WISE-FT) the finetuned models and its pretrained (non finetuned) counterpart \cite{Wortsman_2022_CVPR, WortsmanPatching}. While prior work used this procedure to obtain robustness on out-of-distribution test sets, we use the procedure to improve the robustness of our model when there is a difference between the train and test modalities. 

Denote by $\theta_\textrm{masd}$ be the weights obtained by MASD and $\theta_\textrm{lp}$ be the weights obtained via linear probing. We compute our new weights by taking a weighted average:
\begin{equation}
    \theta_{\textrm{wise}} = \alpha \theta_\textrm{masd} + (1 - \alpha) \theta_\textrm{lp}
\end{equation}

The only deviation from \citet{Wortsman_2022_CVPR} is that they averaged the finetuned image network with the pretrained network weights and ``zero-shot" weights induced by the text embeddings of the class names. Since we finetune all the encoders and want a procedure that is modality agnostic we replace the text based zero-shot weights with linear probe weights. While the choice of $\alpha$ can be tuned with cross-validation we find a constant value of $\alpha=0.75$ works well for our experiments.

\section{Experimental Setup}\label{sec:experiments}

We provide a brief summary of the experimental setup. For complete details, see Appendix. 

{\bf AudioSet}~\cite{gemmeke2017audio} is a video, audio, and text multi-label audio classification dataset over 527 classes. 
Prior work has largely leveraged the audio and/or video, but we also include the title of the video as text. AudioSet consists of an unbalanced training set of 1,743,790 examples, used as unlabeled pretraining data; a training and evaluation sets of 18,649 and 17,065 examples respectively used for the downstream task.

Note that the title is related to the content but rarely contains the audio event label (in $25.5\%$ of the training video titles we have the label word mentioned; for examples see Table~\ref{tab:audioset-examples}). 

{\bf Kinetics-400}~\cite{kay2017kinetics} is a video and audio action recognition dataset over 400 classes. It consists of a training and evaluation sets of 246,245 and 40,000 examples respectively used for the downstream task.

{\bf ImageNet-Captions}~\cite{imagenet-captions} is an image-text dataset created by extracting Flickr captions for images from the original ILSVRC2012 training dataset. It contains 999/1000 of the original ImageNet classes. The dataset contains 448,896 examples which we randomly split into 359,116 training and 89,779 evaluation images. 



{\bf Preprocessing}. We employ standard preprocessing before inference and training for each modality (e.g. \citep{gong2021ast, nagrani2021attention}). Briefly, audio is extracted as single-channel $8$ sec snippet sampled at 16 kHz with necessary padding. We compute log Mel spectrograms (128 frequency bins, 25ms Hamming window, 10 ms stride), and extract $16\times16$ patches.
During training, videos are randomly short-side rescaled between 256 and 320 pixels, and randomly cropped to $224\times224$. During inference, videos are fixed short-side rescaled to 256 pixels following by a center crop to $224\times224$.



{\bf Training}. We use a ViT-B/16 architecture~\cite{dosovitskiy2020image} for all three modalities with appropriate modality specific positional encodings for both contrastive learning and MAE. We initialize  weights for the contrastive model with CLIP ViT-B/16~\cite{radford2021clip}. For AudioSet and Kinetics-400, we learn multimodal representation using AudioSet 2M. For ImageNet-Captions we use the OpenAI released ViT-B-16 CLIP representation.

In the MASD loss in Eq.~\ref{eq:masd_loss} we need a modality complete unlabeled data $D_{SD}$ for self distillation. For experiments on AudioSet $D_{SD}$ is a random 20K sample from the pre-training AudioSet data. For experiments on Kinetics $D_{SD}$ is either a random 20K sample from pre-training AudioSet data or $20\%$ random sample from the Kinetics training data. In the latter case the downstream task training data consists of the remaining $80\%$.


We train models with a 1024 batch size using the AdamW optimizer~\cite{loshchilov2017decoupled} with a learning rate of 8e-4. We pretrain the MAE and contrastive models 256 and 32 epochs, respectively.

\section{Multimodal Robustness Analysis}\label{sec:lessons}
In the following we provide an analysis of multimodal models focusing on the following high level questions:
\setlist{nolistsep}
    \begin{enumerate}[noitemsep,leftmargin=*]
    \item How do different multimodal representation learning methods fare against discrepancies between downstream training and evaluation modalities?
    \item What type of discrepancies have the strongest impact on peformance and/or robustness?
    \item What is the effect of the proposed interventions from Sec.~\ref{sec:interventions} on multimodal robustness?
\end{enumerate}

\subsection{Analysis of multimodal learned representations}\label{sec:common_ssl_analysis}
We focus on learned representations from standard contrastive learning and MAE presented in Sec.~\ref{sec:standard_models}. Performance and Robustness metrics are presented in Tab. \ref{tab:distillation_sliced_results}.


{\bf More modalities are better}. To motivate the use of \textit{multiple modalities during pretraining, training and evaluation} we measure the performance of contrastive learning, the better performing SSL model
during both pretraining and downstream training, while maintaining $M_T = M_E$. We compute Performance per Eq. \ref{eq:metrics} where we average only across modality sets of a fixed size $|M_T|=|M_E|=k$. We vary $k\in\{1,2,3\}$.

\begin{figure}
    \centering
    \includegraphics[width=0.4\textwidth]{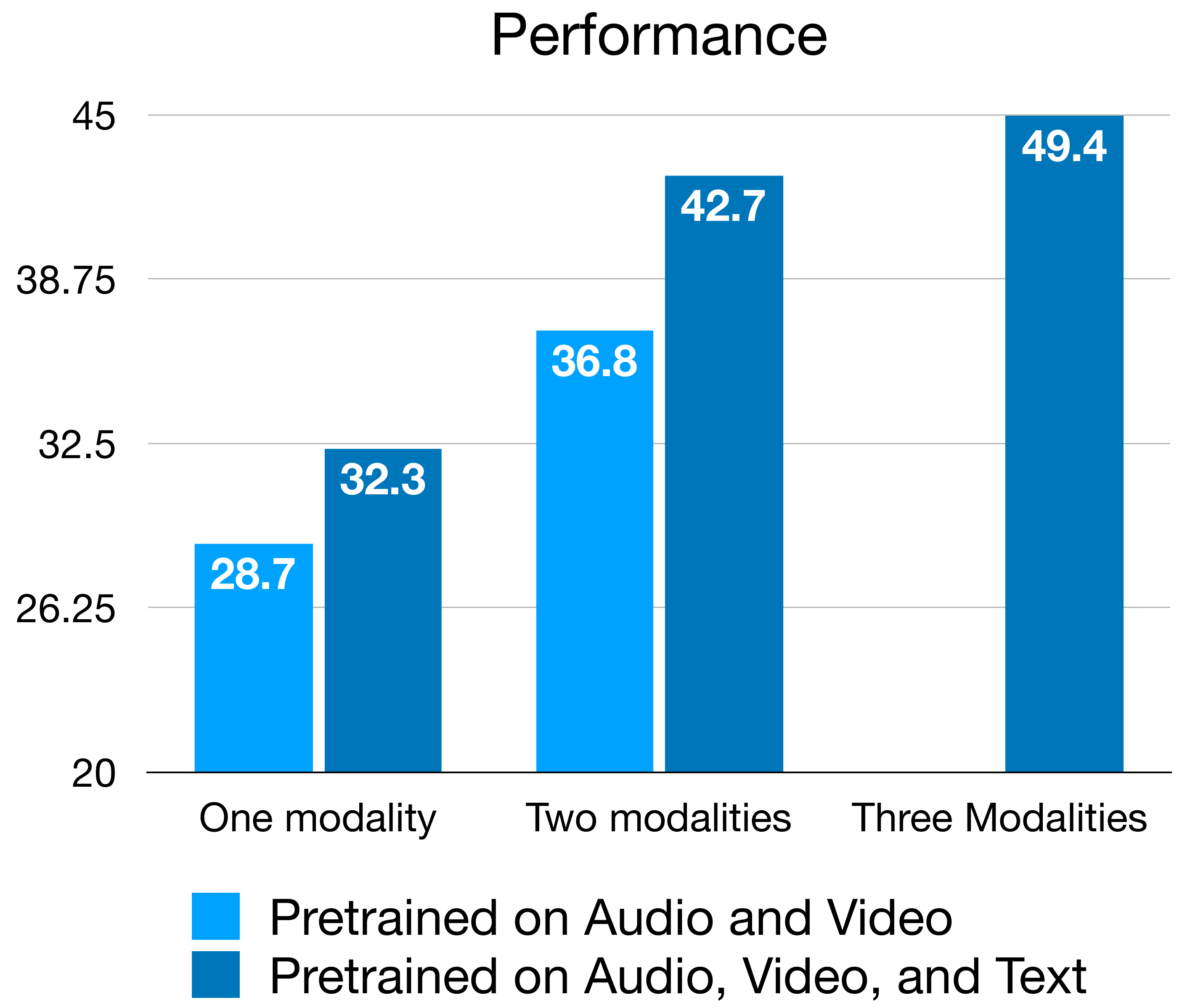}
    \caption{{\bf Increasing the number of modalities at pretraining improves performance.} We consider two models pretrained using Contrastive Learning, one using audio and video, and a second using audio, video, and text. These are applied on a downstream task using 1-3 modalities. The performance numbers are averages across all possible combinations of 1-3 modalities, accordingly. Note, that the only the 3-modality pretrained model can be applied on 3 modalities, hence the right side of the plot has only one model.}
    \label{fig:the-more-the-better}
\end{figure}

Performance consistently improves as a model trains and tests on additional modalities (Fig. \ref{fig:the-more-the-better}).
Furthermore, the models benefit from more modalities at both pretraining and downstream training time. More specifically, 
pretraining on more modalities boosts performance further by 3.5 - 6.0 points (Fig. \ref{fig:the-more-the-better}, light vs dark blue).

{\bf Multimodal representation struggles at downstream task for modalities not seen during training}. The metrics introduced in Tab. \ref{tab:distillation_sliced_results} (\textit{Overall}) aggregate across all possible training and evaluation combinations. To better understand which combinations challenge these models the most, we utilize the startified Performance and Robustness metrics \textit{Added at Test}, \textit{Missing at Test}, and \textit{Transferability} defined in Sec.~\ref{subsec:metrics}. Table~\ref{tab:distillation_sliced_results} (right) shows results over these metrics. 

The first observation is that the models are most robust when we have additional modalities at evaluation. In addition, the gap between robustness and average performance for both SSL methods is quite small in this case, which means that additional modalities during evaluation tend to only improve results. It's worth noting that, since the additional evaluation modalities were not present during downstream training, many of their associated parameters have not changed since pretraining, and yet they can still be combined with the finetuned parameters and improve evaluation performance. This is particular interesting for MAE, since all input modalities must pass through the final modality-shared encoder layer.

In the case of missing modalities at evaluation we see a small performance drop and a large robustness drop for both methods, although the degradation is worse for MAE \footnote{For example, on an AVT trained model, AV performance is $84.8\%$ and $74.3\%$ of AVT for contrastive and MAE, respectively.}. Of course, some performance degradation is expected when modalities are removed. Ideally, performance should degrade \textit{gracefully}, meaning it performs not significantly worse on the evaluation modalities than it would if those were the same modalities used during training. 

In the case of completely different modalities at evaluation, we see that contrastive learning exhibits some transferability properties, but MAE collapses completely. This is again expected due to the difference in pretraining objectives, and since the only modality-shared parameters for contrastive models are the final linear classifier head whereas the MAE encoder also has modality-shared parameters in its final transformer layer. This seems to be the most challenging setup for all SSL methods.

\begin{table*}[th]
\footnotesize
    \centering
    \begin{tabular}{|c|c||c|c|c|c||c|c|c|c|c|c|}
\hline
\multicolumn{12}{|c|}{Contrastive Loss Pretraining Method} \\
\hline
\multirow{2}{*}{\shortstack{Dataset}} &\multirow{2}{*}{\shortstack{Downstream \\ Task Training}} & \multicolumn{4}{c||}{Overall} & \multicolumn{2}{c|}{Missing at Test} & \multicolumn{2}{c|}{Added at Test} &  \multicolumn{2}{|c|}{Transferability}\\
\cline{3-12}
& & $P_\textrm{best}$ & $R_\textrm{best}$ & P & R & P & R & P & R & P & R \\
\hline\hline
    \multirow{7}{*}{AudioSet}  & linear probe & 33.7 &  22.1 & 28.0 &  15.0 &  29.6 &  24.0 &  34.2 &  33.5 & 16.0 &  13.9 \\
    \cline{2-12}
     & fine-tune  &  36.5 &  20.8 & 29.9 &  13.8 &  31.2 &  23.6 &  38.1 &  37.4 & 15.1 &  13.0 \\
    \cline{2-12}
     & WiseFT & 37.3 &  22.3 & 29.5 &  13.5 &  31.4 &  24.5 &  37.6 &  36.9 & 14.0 &  12.0 \\
    \cline{2-12}
       & MASD   &  \textbf{37.4} &  24.1 & 33.5 &  21.9 &  30.5 &  22.4 &  \textbf{40.4} &  39.7 & 26.1 &  24.1\\
    \cline{2-12}
       & MASD+WiseFT   &  37.3 &  \textbf{24.8} & \textbf{33.9} &  \textbf{22.8} &  \textbf{31.3} &  \textbf{24.1} &  40.2 &  \textbf{39.5} & \textbf{26.3} &  \textbf{24.3}\\
    \cline{2-12}
    \cline{2-12}
       & fine-tune on 2M & 37.0 &  21.8 & 32.7 &  18.2 &  30.5 &  23.5 &  41.3 &  40.3 & 20.3 &  18.2 \\
       
\hline\hline
\multirow{4}{*}{\shortstack{Kinetics- \\ 400}}  & linear probe & 42.2 &  21.7 & 34.7 &  17.0 &  34.4 &  \textbf{18.5} &  \multicolumn{2}{c|}{36.8\textsuperscript{*}} & \multicolumn{2}{c|}{16.2\textsuperscript{*}} \\
\cline{2-12}
 & fine-tune  & 45.5 &  11.1 & 36.2 &   6.1 &  29.1 &  11.1 &  \multicolumn{2}{c|}{47.8\textsuperscript{*}} &   \multicolumn{2}{c|}{3.6\textsuperscript{*}} \\
    \cline{2-12}
& MASD, distill-on-AS & 49.8 & 23.5	& 40.6 & 18.5 & \textbf{37.7} & 17.5	& \multicolumn{2}{c|}{49.0\textsuperscript{*}} & \multicolumn{2}{c|}{19.1\textsuperscript{*}}
 \\
\cline{2-12} 
 & MASD, distill-on-Kinetics & \textbf{52.0} & \textbf{26.9} & \textbf{45.2} & \textbf{19.9} & 29.1	& 11.1 & \multicolumn{2}{c|}{\textbf{59.0}\textsuperscript{*}} & \multicolumn{2}{c|}{\textbf{33.7}\textsuperscript{*}}
 \\
 
\hline
\hline
\multirow{3}{*}{\shortstack{ImageNet-\\Captions}}  & linear probe & 70.5 & 68.4 & 66.0 & 48.8 & 70.5 & 68.4 & \multicolumn{2}{c|}{74.3\textsuperscript{*}} & \multicolumn{2}{c|}{39.1\textsuperscript{*}} \\
\cline{2-12} 
 & fine-tune  & 78.7 & 66.7 & 75.4 & 58.7 & 72.0 & 66.7 & \multicolumn{2}{c|}{85.3\textsuperscript{*}} & \multicolumn{2}{c|}{54.7\textsuperscript{*}} \\
   \cline{2-12}
 & MASD &  \textbf{84.3}	& \textbf{80.8} & \textbf{82.4} & \textbf{76.0} & 72.0 & 66.7 & \multicolumn{2}{c|}{\textbf{90.9}\textsuperscript{*}} & \multicolumn{2}{c|}{\textbf{80.8}\textsuperscript{*}} \\
\hline
    \end{tabular}
\newline
\vspace*{0.3cm}
\newline
\begin{tabular}{|c|c||c|c|c|c||c|c|c|c|c|c|}
\hline
\multicolumn{12}{|c|}{Masked Autoencoder Pretraining Method} \\
\hline
\multirow{2}{*}{\shortstack{Dataset}}   &\multirow{2}{*}{\shortstack{Downstream \\ Task Training}} & \multicolumn{4}{c||}{Overall} & \multicolumn{2}{c|}{Missing at Test} & \multicolumn{2}{c|}{Added at Test} &  \multicolumn{2}{c|}{Transferab.}\\
\cline{3-12}
& & $P_\textrm{best}$ & $R_\textrm{best}$ & P & R & P & R & P & R & P & R \\
\hline\hline
\multirow{3}{*}{AudioSet}  & linear probe & 23.4 &   5.5 & 14.0 &  1.8 &  17.2 &   7.7 &  17.8 &  17.0 &   1.5 & 1.2 \\
\cline{2-12}
 & fine-tuned &  28.9 &   3.8 & 20.0 &  1.3 &  21.4 &  10.0 &  30.8 &  30.3 &   1.1 & 0.9
\\
    \cline{2-12}
 & MASD & \textbf{30.6} &  \textbf{15.1} & \textbf{26.6} &  \textbf{9.5} &  \textbf{21.6} &  \textbf{10.3} &  \textbf{35.4} &  \textbf{34.4} &  \textbf{18.5} & \textbf{15.1} \\
 \hline\hline
 
\multirow{4}{*}{\shortstack{Kinetics-\\400}}  & linear probe & 30.6	& 11.0 & 19.8 & 3.9 & 19.6 & 11.0 &  \multicolumn{2}{c|}{7.1\textsuperscript{*}} &  \multicolumn{2}{c|}{0.4\textsuperscript{*}}
 \\
\cline{2-12}
 & fine-tuned &  50.5 & 17.1 & 38.2 & 5.9 & 40.2 & 17.1 & \multicolumn{2}{c|}{47.0\textsuperscript{*}} & \multicolumn{2}{c|}{0.3\textsuperscript{*}} \\
\cline{2-12}
 & MASD, distill-on-AS & 49.0 & 19.2 & 41.4 & 15.7& 38.5 & \textbf{19.2}	& \multicolumn{2}{c|}{50.6\textsuperscript{*}} & \multicolumn{2}{c|}{14.0\textsuperscript{*}} \\
\cline{2-12} 
 & MASD, distill-on-Kinetics & \textbf{53.1}	& \textbf{27.6}	& \textbf{49.1} & \textbf{21.5} & 40.2 & 17.1 & \multicolumn{2}{c|}{\textbf{61.9}\textsuperscript{*}} & \multicolumn{2}{c|}{\textbf{34.7}\textsuperscript{*}} \\

\hline
    \end{tabular}

    \caption{Best Performance ($P_\textrm{best}$), Best Robustness ($R_\textrm{best}$), Average Performance (P) and Robustness (R) for two pretraining techniques with and without MASD, WiseFT: \textbf{top} is Contrastive Learning, \textbf{bottom} is Mask Autoenconder. We show results on AudioSet using audio, video, and text; Kinetics-400 using audio and video; and ImageNet-Captions with image and text. On the left side under \textit{Overall} we show metrics computed over all possible training/evaluation modalities, on the right we show results for specific training/evaluation modality combinations (see Sec.~\ref{sec:common_ssl_analysis}). For Kinetics and AudioSet experiments we pretrain on AudioSet only. During self-distillation on Kinetics, we provide experiments by using AudioSet or a held-out portion of Kinetics.\textsuperscript{*} For datasets with two modalities, per Sec.~\ref{sec:lessons}, the values for robustness and performance for these training/evaluation combinations are identical.}
    \label{tab:distillation_sliced_results}
\end{table*}

\begin{figure}[t]
    \centering
    \includegraphics[width=0.45\textwidth]{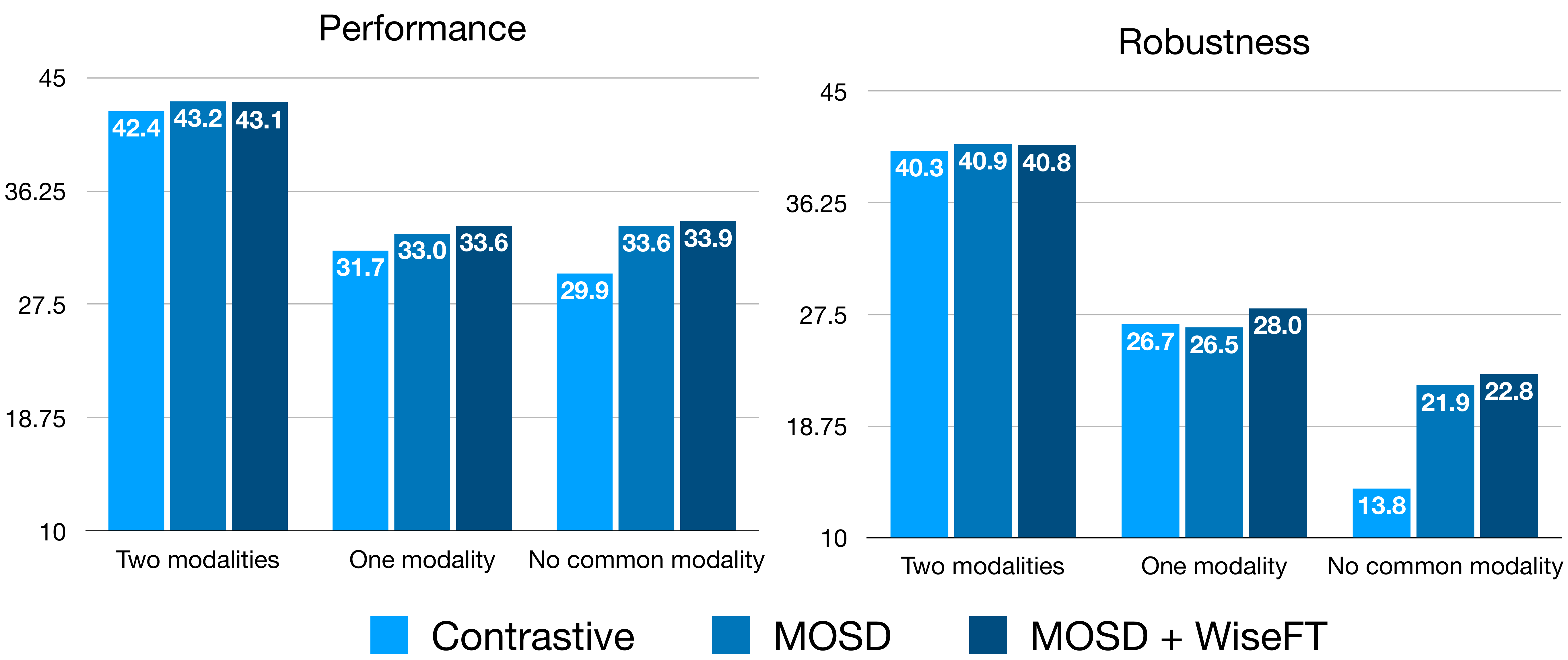}
    \vspace{-0.4cm}
    \caption{{\bf MASD and WiseFT improves performance and robustness.} Average Performance (P) and Robustness (R) as the number of overlapping modalities between training/test goes from Two, to One, to None, for Contrastive Learning, Contrastive + MASD, and Contrastive + MASD + WiseFT.}
    \label{fig:performance_per_overlapping_modalities_us_vs_them}
\end{figure}

\begin{table}[t]
\tiny
    \centering
    \begin{tabular}{|c|c|c|c|c|c|c|c|}
\hline
\multirow{2}{*}{\shortstack{SSL \\ Pretraining}}  &\multirow{2}{*}{\shortstack{Downstream \\ Task Training}} & \multicolumn{6}{c|}{Downstream Task Training Modalities} \\
\cline{3-8}
& & V  &  A  &    T  &   AV  &  AT  &   VT \\
    \hline\hline
    Contrastive & fine-tune  &  AVT  &    A  &  AVT  &  AVT  &   AT  &  AVT  \\
\hline
      Contrastive & MASD   &  AVT  &  AVT  &  AVT  &  AVT  &  AVT  &  AVT \\
\hline
      Contrastive & MASD+WiseFT   &  AVT  &  AVT  &  AVT  &  AVT  &  AVT  &  AVT \\
\hline
\hline
MAE & fine-tune & VT  &  AT  &    T  &   AVT &   AT  &   VT  \\
\hline
MAE & MASD & AVT  &  AVT  &  AVT  &  AVT  &  AVT  &  AVT \\
\hline

    \end{tabular}
    \caption{For each training modality set, we show the combination of evaluation modalities yielding the highest performance (see text). We abbreviate video=V, audio=A, text=T.}
    \label{tab:best_testing_modalities}
\end{table}

\begin{table}
\tiny
    \centering
    \begin{tabular}{|c|c|c|c|c|c|}
    \hline
       \multirow{2}{*}{Model}  &  \multirow{2}{*}{Pretrain} & \multicolumn{4}{c|}{Training/Evaluation Modalities} \\
       \cline{3-6} 
        & &  A& V & AV & AVT\\
       \hline\hline
       Contrastive, FT  & AS2M &  39.5& 25.6 & 43.7 & 49.4 \\
       \hline
       Contrastive, MASD+WiseFT  & AS2M &  39.5& 30.0 & 44.2 & 49.4 \\
       \hline
        MBT~\citep{nagrani2021attention} & IN21K &  31.3 & 27.7 & 43.9 & \\
        \hline
        CAV-MAE~\cite{gong2022contrastive} & AS2M &  37.7 & 19.8 & 42.0 & \\
        \hline
        VATT~\cite{akbari2021vatt} & IN & 39.4 &  &  &  \\
        \hline
        Audio-MAE~\cite{huang2022audiomae} & AS2M &  37.0 &  &  &  \\
        \hline
    \end{tabular}
    \caption{Mean Average Precision on AudioSet 20K test for standard Contrastive Learning, MASD, and other competitive approaches on AudioSet. For pre-training, one can used either AS2M, ImageNet, or ImageNet 21K~\cite{deng2009imagenet}. Results not present in the literature are empty.}
    \label{tab:comparison}
\end{table}

\subsection{Analysis of Robustness Interventions}
The performance and robustness of propsed interventions from Sec.~\ref{sec:interventions} and baseline models are shown in Table~\ref{tab:distillation_sliced_results}. These metrics are presented as an average across all training/evaluation modality combinations as well as across combination slices identified in Sec.~\ref{sec:analysis}. 

{\bf MASD improve both performance and robustness}
As a first observation, MASD leads to a Performance improvement and substantial improvement of Robustness, for  AudioSet, Kinetics-400, and ImageNet-Captions. Thus, MASD is addressing the weaknesses of original SSL methods. In particular, it reduced the degradation in case of \textit{Added at Test} and \textit{Transferability}, and in the case of Contrastive Learning, MASD doubles both Performance and Robustness. These results are consistent across both datasets, which demonstrates the generality of the learnings. The only degradation is in \textit{Missing at Test} which is fixed by Wise-FT.
Furthermore, our results show that MASD generalizes across three different types of modality sets across AudioSet, Kinetics-400, and ImageNet-Captions.

\begin{figure}
\centering
\includegraphics[width=0.45\textwidth]{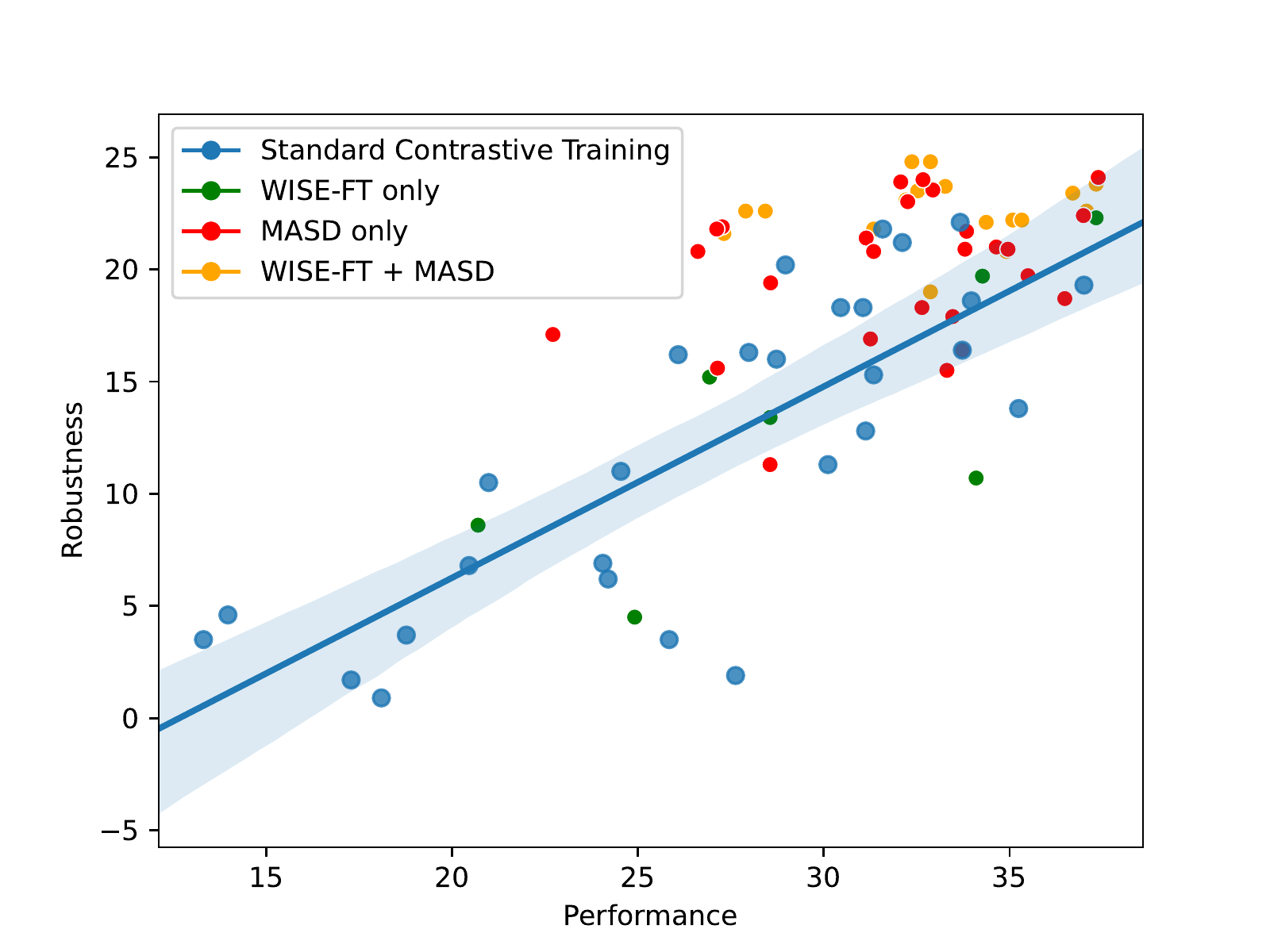}
\caption{{\bf Interventions improve robustness.} WiSE-FT + MASD provide substantial improvements to robustness across most training modality sets $M_T$. For each of the four methods and each possible training modality set $M_T\subseteq\{\textrm{audio}, \textrm{video}, \textrm{text}\}$, we plot robustness vs performance per Eq.~(\ref{eq:robustness_per_training_mods}) and (\ref{eq:performance_per_training_mods}).}
\label{fig:perf_robust}
\vspace{-0.3cm}
\end{figure}

To further see the benefit of our proposed interventions we plot Robustness vs Performance for each possible training modality set in Fig.~\ref{fig:perf_robust}. While we see that Robustness is generally correlated with Performance, our interventions when combined consistently improve Robustness beyond the trend line. This is similar to a notion of ``high effective robustness" as defined in \cite{taori2020}.

{\bf MASD improves robustness beyond supervised learning on more examples} A natural question is whether downstream supervised training on larger labeled data can address multimodal robustness issues. In Table~\ref{tab:distillation_sliced_results}, we present downstream training on 2M labeled examples, which is $100\times$ than the labeled downstream training data for all other experiments. Although we see a $50\%$ boost in robustness compared to regular downstream fine-tuning, this experiments still underperforms MASD on Robustness, in particular for \textit{Transferability}, while using substantially more labeling.

{\bf Robustness gains correlate with train-test modality gap.} To better understand MASD, we compute metrics as we decrease the number of common modalities between training and evaluation. In Fig.~\ref{fig:performance_per_overlapping_modalities_us_vs_them}, we show Performance and Robustness for $k = |M_T\cap M_E| \in\{0,1, 2\}$ (see Eq.~(\ref{eq:metrics})), i.e.~zero, one, or two common modalities. We can see that as the number of common modalities decreases, MASD degrades more gracefully compared to standard Contrastive Learning. WiseFT provides an additional stability in performance.

{\bf MASD helps better utilize all modalities at evaluation time.} Another property of MASD is that it can utilize all modalities present at downstream evaluation, even if these are not available at downstream training. To see this, for each downstream training modality set $M_T$ we identify the evaluation modality set $M_E$ yielding highest performance: $\arg\max_{M_E\subseteq M} p(M_E; M_T) \textrm{ for each } M_T\subseteq M$.

We summarize the best evaluation modalities for each training modality set in Table~\ref{tab:best_testing_modalities}. We can see that for the original Contrastive learning, in $2$ out of $6$ training setups the model attains best performance using the same evaluation modalities it has been trained on, $M_E = M_T$. However, for MASD we see that it always works best when we use all modalities, $M_E=\{A, V, T\}$. For MAE, we see an even bigger utilization -- while in $5$ cases the original model prefers a subset of the modalities at evaluation, with MASD the model in all $6$ cases benefits from having all modalities at evaluation. 

{\bf MASD achieves competitive performance compared to other approaches} To better put MASD in perspective, we compare its performance to other approaches in the literature. In Table~\ref{tab:comparison}, we show results using the same training and evaluation modalities, we do so for four different modality sets: audio only, video only; audio and video; audio, video, and text. When using AudioSet 20K downstream training set as only labeled data, MASD achieves higher or equal performance to other reported approaches, across all studied modality combinations. Further, if using text, we obtain even superior performance (although other approaches do not use text). This shows that MASD not only fixes robustness issues for underlying SSL methods, but also keeps competitive results across various evaluation setups. We note that our AV number is the best reported number among all methods that only have access to the AS-20k labels. 

\vspace{-0.2cm}
\section{Discussion}
In this paper we quantified the notion of robustness in a multimodal representation. We introduced several simple definitions of robustness based on average and worst case performance across subsets of modalities. We characterized the robustness of state-of-the-art learned representations based on contrastive learning and masked autoencoders.

We found that performance degrades with greater discrepancies between training and testing modalities, however these degradations may be alleviated with training improvements based on MASD distillation and WiseFT aggregation. Using these techniques we are able to improve upon state-of-the-art with AudioSet by leveraging multimodal data not available to the downstream task.

We observe several limitations for this current work, and opportunities for extensions and next steps.
First, we focused our representation learning on homogenous multimodal data and it is unclear how this work will succeed in large scale hetergenous datasets.
Further, although our benchmarks quantify the multimodal behavior on several datasets, it is unclear what is truly achievable given the structure and features of a given dataset. We strongly suspect that these results may be heavily dependent on the specifics of a given multimodal dataset but much work remains to characterize how the trends identified persist and how these benchmarks vary across typical multimodal conditions.
\section{Author Contributions and Acknowlegements}

{\bf Brandon McKinzie} implemented majority of codebase; drove research directions; improved upon initial designs for various model architectures and training objectives; assisted in formulating metrics; ran all of experiments except wise-ft, and initial MAE experiments; wrote appendix and helped with main paper writing.

{\bf Vaishaal Shankar} co-scoped the main metrics of interest for the paper; proposed and ran all the WISE-FT experiments; proposed, defined and ran all the ImageNet captions experiments; wrote initial version of introduction and co-wrote related work sections; helped with main paper writing.

{\bf Joseph Cheng} helped set up the codebase; implemented audio preprocessing; help implement video inputs, implemented MAE; ran initial experiments on MAE and AudioSet; wrote related work section

{\bf Jonathon Shlens} advised on the project, discussed experiments, assisted with the analysis, and helped on the writing.

{\bf Yinfei Yang} advised on the project, provided feedback on writing.

{\bf Alex Toshev} initiated the project, led research direction, co-designed the robustness evaluation framework; designed the main algorithmic contributions of the paper; wrote most of the paper.

The authors would like to thank Jason Ramapuram and Tatiana Likhomanenko for useful suggestions regarding Knowledge Distillation; Jason Ramapuram, Devon Hjelm, Hadi Pour Ansari, and Barry Theobold for detailed feedback on the experiments, algorithm design, overall paper structure and writing; Oncel Tuzel, Sachin Mehta, Fartash Faghri, Alkesh Patel for ongoing feedback during the project; Tom Nickson and Angelos Katharopoulos for ongoing infrastructure support.

\bibliography{main}

\begin{thebibliography}{69}
\providecommand{\natexlab}[1]{#1}
\providecommand{\url}[1]{\texttt{#1}}
\expandafter\ifx\csname urlstyle\endcsname\relax
  \providecommand{\doi}[1]{doi: #1}\else
  \providecommand{\doi}{doi: \begingroup \urlstyle{rm}\Url}\fi

\bibitem[Akbari et~al.(2021)Akbari, Yuan, Qian, Chuang, Chang, Cui, and
  Gong]{akbari2021vatt}
Akbari, H., Yuan, L., Qian, R., Chuang, W.-H., Chang, S.-F., Cui, Y., and Gong,
  B.
\newblock Vatt: Transformers for multimodal self-supervised learning from raw
  video, audio and text.
\newblock \emph{Advances in Neural Information Processing Systems},
  34:\penalty0 24206--24221, 2021.

\bibitem[Alayrac et~al.(2020)Alayrac, Recasens, Schneider, Arandjelovic,
  Ramapuram, Fauw, Smaira, Dieleman, and Zisserman]{alayrac2020versatile}
Alayrac, J., Recasens, A., Schneider, R., Arandjelovic, R., Ramapuram, J.,
  Fauw, J.~D., Smaira, L., Dieleman, S., and Zisserman, A.
\newblock Self-supervised multimodal versatile networks.
\newblock \emph{CoRR}, abs/2006.16228, 2020.
\newblock URL \url{https://arxiv.org/abs/2006.16228}.

\bibitem[Alayrac et~al.(2022)Alayrac, Donahue, Luc, Miech, Barr, Hasson, Lenc,
  Mensch, Millican, Reynolds, et~al.]{alayrac2022flamingo}
Alayrac, J.-B., Donahue, J., Luc, P., Miech, A., Barr, I., Hasson, Y., Lenc,
  K., Mensch, A., Millican, K., Reynolds, M., et~al.
\newblock Flamingo: a visual language model for few-shot learning.
\newblock \emph{arXiv preprint arXiv:2204.14198}, 2022.

\bibitem[Alzantot et~al.(2018)Alzantot, Sharma, Elgohary, Ho, Srivastava, and
  Chang]{alzantot2018generating}
Alzantot, M., Sharma, Y., Elgohary, A., Ho, B.-J., Srivastava, M., and Chang,
  K.-W.
\newblock Generating natural language adversarial examples.
\newblock \emph{arXiv preprint arXiv:1804.07998}, 2018.

\bibitem[Baade et~al.(2022)Baade, Peng, and Harwath]{baade2022mae}
Baade, A., Peng, P., and Harwath, D.
\newblock Mae-ast: Masked autoencoding audio spectrogram transformer.
\newblock \emph{arXiv preprint arXiv:2203.16691}, 2022.

\bibitem[Barbu et~al.(2019)Barbu, Mayo, Alverio, Luo, Wang, Gutfreund,
  Tenenbaum, and Katz]{barbu2019objectnet}
Barbu, A., Mayo, D., Alverio, J., Luo, W., Wang, C., Gutfreund, D., Tenenbaum,
  J., and Katz, B.
\newblock Objectnet: A large-scale bias-controlled dataset for pushing the
  limits of object recognition models.
\newblock \emph{Advances in neural information processing systems}, 32, 2019.

\bibitem[Boser et~al.(1992)Boser, Guyon, and Vapnik]{boser1992training}
Boser, B.~E., Guyon, I.~M., and Vapnik, V.~N.
\newblock A training algorithm for optimal margin classifiers.
\newblock In \emph{Proceedings of the fifth annual workshop on Computational
  learning theory}, pp.\  144--152, 1992.

\bibitem[Carlini \& Wagner(2018)Carlini and Wagner]{carlini2018audio}
Carlini, N. and Wagner, D.
\newblock Audio adversarial examples: Targeted attacks on speech-to-text.
\newblock In \emph{2018 IEEE security and privacy workshops (SPW)}, pp.\  1--7.
  IEEE, 2018.

\bibitem[Carlini et~al.(2022)Carlini, Tramer, Dvijotham, and
  Kolter]{carliniAdvCert}
Carlini, N., Tramer, F., Dvijotham, K., and Kolter, J.~Z.
\newblock (certified!!) adversarial robustness for free!, 2022.
\newblock URL \url{https://arxiv.org/abs/2206.10550}.

\bibitem[Carmon et~al.(2019)Carmon, Raghunathan, Schmidt, Duchi, and
  Liang]{carmonDefense}
Carmon, Y., Raghunathan, A., Schmidt, L., Duchi, J.~C., and Liang, P.~S.
\newblock Unlabeled data improves adversarial robustness.
\newblock In Wallach, H., Larochelle, H., Beygelzimer, A., d\textquotesingle
  Alch\'{e}-Buc, F., Fox, E., and Garnett, R. (eds.), \emph{Advances in Neural
  Information Processing Systems}, volume~32. Curran Associates, Inc., 2019.
\newblock URL
  \url{https://proceedings.neurips.cc/paper/2019/file/32e0bd1497aa43e02a42f47d9d6515ad-Paper.pdf}.

\bibitem[Castro et~al.(2018)Castro, Mar{\'\i}n-Jim{\'e}nez, Guil, Schmid, and
  Alahari]{castro2018end}
Castro, F.~M., Mar{\'\i}n-Jim{\'e}nez, M.~J., Guil, N., Schmid, C., and
  Alahari, K.
\newblock End-to-end incremental learning.
\newblock In \emph{Proceedings of the European conference on computer vision
  (ECCV)}, pp.\  233--248, 2018.

\bibitem[Chong et~al.(2022)Chong, Wang, Zhou, and Zeng]{chong2022maskspec}
Chong, D., Wang, H., Zhou, P., and Zeng, Q.
\newblock Masked spectrogram prediction for self-supervised audio pre-training,
  2022.
\newblock URL \url{https://arxiv.org/abs/2204.12768}.

\bibitem[Deng et~al.(2009)Deng, Dong, Socher, Li, Li, and
  Fei-Fei]{deng2009imagenet}
Deng, J., Dong, W., Socher, R., Li, L.-J., Li, K., and Fei-Fei, L.
\newblock Imagenet: A large-scale hierarchical image database.
\newblock In \emph{CVPR}, pp.\  248--255. Ieee, 2009.

\bibitem[Devlin et~al.(2018)Devlin, Chang, Lee, and Toutanova]{devlin2018bert}
Devlin, J., Chang, M.-W., Lee, K., and Toutanova, K.
\newblock Bert: Pre-training of deep bidirectional transformers for language
  understanding.
\newblock \emph{arXiv preprint arXiv:1810.04805}, 2018.

\bibitem[Dosovitskiy et~al.(2020)Dosovitskiy, Beyer, Kolesnikov, Weissenborn,
  Zhai, Unterthiner, Dehghani, Minderer, Heigold, Gelly,
  et~al.]{dosovitskiy2020image}
Dosovitskiy, A., Beyer, L., Kolesnikov, A., Weissenborn, D., Zhai, X.,
  Unterthiner, T., Dehghani, M., Minderer, M., Heigold, G., Gelly, S., et~al.
\newblock An image is worth 16x16 words: Transformers for image recognition at
  scale.
\newblock \emph{arXiv preprint arXiv:2010.11929}, 2020.

\bibitem[Dou et~al.(2021)Dou, Xu, Gan, Wang, Wang, Wang, Zhu, Zhang, Yuan,
  Peng, Liu, and Zeng]{dou2021empiricalvisionandlanguage}
Dou, Z.-Y., Xu, Y., Gan, Z., Wang, J., Wang, S., Wang, L., Zhu, C., Zhang, P.,
  Yuan, L., Peng, N., Liu, Z., and Zeng, M.
\newblock An empirical study of training end-to-end vision-and-language
  transformers, 2021.
\newblock URL \url{https://arxiv.org/abs/2111.02387}.

\bibitem[Fang et~al.(2022{\natexlab{a}})Fang, Ilharco, Wortsman, Wan, Shankar,
  Dave, and Schmidt]{imagenet-captions}
Fang, A., Ilharco, G., Wortsman, M., Wan, Y., Shankar, V., Dave, A., and
  Schmidt, L.
\newblock Data determines distributional robustness in contrastive language
  image pre-training ({CLIP}).
\newblock In Chaudhuri, K., Jegelka, S., Song, L., Szepesvari, C., Niu, G., and
  Sabato, S. (eds.), \emph{Proceedings of the 39th International Conference on
  Machine Learning}, volume 162 of \emph{Proceedings of Machine Learning
  Research}, pp.\  6216--6234. PMLR, 17--23 Jul 2022{\natexlab{a}}.
\newblock URL \url{https://proceedings.mlr.press/v162/fang22a.html}.

\bibitem[Fang et~al.(2022{\natexlab{b}})Fang, Wang, Xie, Sun, Wu, Wang, Huang,
  Wang, and Cao]{fang2022eva}
Fang, Y., Wang, W., Xie, B., Sun, Q., Wu, L., Wang, X., Huang, T., Wang, X.,
  and Cao, Y.
\newblock Eva: Exploring the limits of masked visual representation learning at
  scale, 2022{\natexlab{b}}.
\newblock URL \url{https://arxiv.org/abs/2211.07636}.

\bibitem[Feichtenhofer et~al.(2022)Feichtenhofer, Fan, Li, and
  He]{feichtenhofer2022masked}
Feichtenhofer, C., Fan, H., Li, Y., and He, K.
\newblock Masked autoencoders as spatiotemporal learners.
\newblock \emph{arXiv preprint arXiv:2205.09113}, 2022.

\bibitem[Geirhos et~al.(2018)Geirhos, Temme, Rauber, Sch\"{u}tt, Bethge, and
  Wichmann]{Geirhos2018}
Geirhos, R., Temme, C. R.~M., Rauber, J., Sch\"{u}tt, H.~H., Bethge, M., and
  Wichmann, F.~A.
\newblock Generalisation in humans and deep neural networks.
\newblock In Bengio, S., Wallach, H., Larochelle, H., Grauman, K.,
  Cesa-Bianchi, N., and Garnett, R. (eds.), \emph{Advances in Neural
  Information Processing Systems}, volume~31. Curran Associates, Inc., 2018.
\newblock URL
  \url{https://proceedings.neurips.cc/paper/2018/file/0937fb5864ed06ffb59ae5f9b5ed67a9-Paper.pdf}.

\bibitem[Gemmeke et~al.(2017)Gemmeke, Ellis, Freedman, Jansen, Lawrence, Moore,
  Plakal, and Ritter]{gemmeke2017audio}
Gemmeke, J.~F., Ellis, D.~P., Freedman, D., Jansen, A., Lawrence, W., Moore,
  R.~C., Plakal, M., and Ritter, M.
\newblock Audio set: An ontology and human-labeled dataset for audio events.
\newblock In \emph{2017 IEEE international conference on acoustics, speech and
  signal processing (ICASSP)}, pp.\  776--780. IEEE, 2017.

\bibitem[Geng et~al.(2022)Geng, Liu, Lee, Schuurams, Levine, and
  Abbeel]{geng2022multimodal}
Geng, X., Liu, H., Lee, L., Schuurams, D., Levine, S., and Abbeel, P.
\newblock Multimodal masked autoencoders learn transferable representations.
\newblock \emph{arXiv preprint arXiv:2205.14204}, 2022.

\bibitem[Georgescu et~al.(2022)Georgescu, Fonseca, Ionescu, Lucic, Schmid, and
  Arnab]{georgescu2022audiovisual}
Georgescu, M.-I., Fonseca, E., Ionescu, R.~T., Lucic, M., Schmid, C., and
  Arnab, A.
\newblock Audiovisual masked autoencoders.
\newblock \emph{arXiv preprint arXiv:2212.05922}, 2022.

\bibitem[Girdhar et~al.(2022)Girdhar, El-Nouby, Singh, Alwala, Joulin, and
  Misra]{girdhar2022omnimae}
Girdhar, R., El-Nouby, A., Singh, M., Alwala, K.~V., Joulin, A., and Misra, I.
\newblock Omnimae: Single model masked pretraining on images and videos.
\newblock \emph{arXiv preprint arXiv:2206.08356}, 2022.

\bibitem[Gong et~al.(2021)Gong, Chung, and Glass]{gong2021ast}
Gong, Y., Chung, Y.-A., and Glass, J.
\newblock Ast: Audio spectrogram transformer.
\newblock \emph{arXiv preprint arXiv:2104.01778}, 2021.

\bibitem[Gong et~al.(2022)Gong, Rouditchenko, Liu, Harwath, Karlinsky, Kuehne,
  and Glass]{gong2022contrastive}
Gong, Y., Rouditchenko, A., Liu, A.~H., Harwath, D., Karlinsky, L., Kuehne, H.,
  and Glass, J.
\newblock Contrastive audio-visual masked autoencoder.
\newblock \emph{arXiv preprint arXiv:2210.07839}, 2022.

\bibitem[Goodfellow et~al.(2016)Goodfellow, Bengio, Courville, and
  Bengio]{goodfellow2016deep}
Goodfellow, I., Bengio, Y., Courville, A., and Bengio, Y.
\newblock \emph{Deep learning}, volume~1.
\newblock MIT Press, 2016.

\bibitem[Gu et~al.(2019)Gu, Yang, Ngiam, Le, and Shlens]{gu2019}
Gu, K., Yang, B., Ngiam, J., Le, Q., and Shlens, J.
\newblock Using videos to evaluate image model robustness, 2019.
\newblock URL \url{https://arxiv.org/abs/1904.10076}.

\bibitem[Gutmann \& Hyv{\"a}rinen(2010)Gutmann and
  Hyv{\"a}rinen]{gutmann2010noise}
Gutmann, M. and Hyv{\"a}rinen, A.
\newblock Noise-contrastive estimation: A new estimation principle for
  unnormalized statistical models.
\newblock In \emph{Proceedings of the thirteenth international conference on
  artificial intelligence and statistics}, pp.\  297--304. JMLR Workshop and
  Conference Proceedings, 2010.

\bibitem[He et~al.(2022)He, Chen, Xie, Li, Doll{\'a}r, and
  Girshick]{he2022masked}
He, K., Chen, X., Xie, S., Li, Y., Doll{\'a}r, P., and Girshick, R.
\newblock Masked autoencoders are scalable vision learners.
\newblock In \emph{Proceedings of the IEEE/CVF Conference on Computer Vision
  and Pattern Recognition}, pp.\  16000--16009, 2022.

\bibitem[Hendrycks \& Dietterich(2019)Hendrycks and
  Dietterich]{hendrycksCorruption}
Hendrycks, D. and Dietterich, T.
\newblock Benchmarking neural network robustness to common corruptions and
  perturbations, 2019.
\newblock URL \url{https://arxiv.org/abs/1903.12261}.

\bibitem[Hessel \& Lee(2020)Hessel and Lee]{hessel2020does}
Hessel, J. and Lee, L.
\newblock Does my multimodal model learn cross-modal interactions? it's harder
  to tell than you might think!
\newblock \emph{arXiv preprint arXiv:2010.06572}, 2020.

\bibitem[Hinton et~al.(2015)Hinton, Vinyals, Dean,
  et~al.]{hinton2015distilling}
Hinton, G., Vinyals, O., Dean, J., et~al.
\newblock Distilling the knowledge in a neural network.
\newblock \emph{arXiv preprint arXiv:1503.02531}, 2\penalty0 (7), 2015.

\bibitem[Hou et~al.(2018)Hou, Pan, Loy, Wang, and Lin]{hou2018lifelong}
Hou, S., Pan, X., Loy, C.~C., Wang, Z., and Lin, D.
\newblock Lifelong learning via progressive distillation and retrospection.
\newblock In \emph{Proceedings of the European Conference on Computer Vision
  (ECCV)}, pp.\  437--452, 2018.

\bibitem[Huang et~al.(2016)Huang, Sun, Liu, Sedra, and
  Weinberger]{huang2016droppath}
Huang, G., Sun, Y., Liu, Z., Sedra, D., and Weinberger, K.
\newblock Deep networks with stochastic depth, 2016.
\newblock URL \url{https://arxiv.org/abs/1603.09382}.

\bibitem[Huang et~al.(2022)Huang, Xu, Li, Baevski, Auli, Galuba, Metze, and
  Feichtenhofer]{huang2022audiomae}
Huang, P.-Y., Xu, H., Li, J., Baevski, A., Auli, M., Galuba, W., Metze, F., and
  Feichtenhofer, C.
\newblock Masked autoencoders that listen, 2022.
\newblock URL \url{https://arxiv.org/abs/2207.06405}.

\bibitem[Ilharco et~al.(2022)Ilharco, Wortsman, Gadre, Song, Hajishirzi,
  Kornblith, Farhadi, and Schmidt]{WortsmanPatching}
Ilharco, G., Wortsman, M., Gadre, S.~Y., Song, S., Hajishirzi, H., Kornblith,
  S., Farhadi, A., and Schmidt, L.
\newblock Patching open-vocabulary models by interpolating weights, 2022.
\newblock URL \url{https://arxiv.org/abs/2208.05592}.

\bibitem[Jaegle et~al.(2021)Jaegle, Borgeaud, Alayrac, Doersch, Ionescu, Ding,
  Koppula, Zoran, Brock, Shelhamer, et~al.]{jaegle2021perceiver}
Jaegle, A., Borgeaud, S., Alayrac, J.-B., Doersch, C., Ionescu, C., Ding, D.,
  Koppula, S., Zoran, D., Brock, A., Shelhamer, E., et~al.
\newblock Perceiver io: A general architecture for structured inputs \&
  outputs.
\newblock \emph{arXiv preprint arXiv:2107.14795}, 2021.

\bibitem[Jia et~al.(2021)Jia, Yang, Xia, Chen, Parekh, Pham, Le, Sung, Li, and
  Duerig]{jia2021scaling}
Jia, C., Yang, Y., Xia, Y., Chen, Y.-T., Parekh, Z., Pham, H., Le, Q.~V., Sung,
  Y., Li, Z., and Duerig, T.
\newblock Scaling up visual and vision-language representation learning with
  noisy text supervision.
\newblock \emph{ICML}, 2021.

\bibitem[Kay et~al.(2017)Kay, Carreira, Simonyan, Zhang, Hillier,
  Vijayanarasimhan, Viola, Green, Back, Natsev, et~al.]{kay2017kinetics}
Kay, W., Carreira, J., Simonyan, K., Zhang, B., Hillier, C., Vijayanarasimhan,
  S., Viola, F., Green, T., Back, T., Natsev, P., et~al.
\newblock The kinetics human action video dataset.
\newblock \emph{arXiv preprint arXiv:1705.06950}, 2017.

\bibitem[Li \& Hoiem(2017)Li and Hoiem]{li2017learning}
Li, Z. and Hoiem, D.
\newblock Learning without forgetting.
\newblock \emph{IEEE transactions on pattern analysis and machine
  intelligence}, 40\penalty0 (12):\penalty0 2935--2947, 2017.

\bibitem[Liang et~al.(2022)Liang, Lyu, Fan, Mo, Yogatama, Morency, and
  Salakhutdinov]{liang2022highmmt}
Liang, P.~P., Lyu, Y., Fan, X., Mo, S., Yogatama, D., Morency, L.-P., and
  Salakhutdinov, R.
\newblock Highmmt: Towards modality and task generalization for high-modality
  representation learning.
\newblock \emph{arXiv preprint arXiv:2203.01311}, 2022.

\bibitem[Liu et~al.(2018)Liu, Li, Xu, and Natarajan]{liu2018learntocombine}
Liu, K., Li, Y., Xu, N., and Natarajan, P.
\newblock Learn to combine modalities in multimodal deep learning, 2018.
\newblock URL \url{https://arxiv.org/abs/1805.11730}.

\bibitem[Loshchilov \& Hutter(2017)Loshchilov and
  Hutter]{loshchilov2017decoupled}
Loshchilov, I. and Hutter, F.
\newblock Decoupled weight decay regularization.
\newblock \emph{arXiv preprint arXiv:1711.05101}, 2017.

\bibitem[Madry et~al.(2017)Madry, Makelov, Schmidt, Tsipras, and
  Vladu]{madry2017towards}
Madry, A., Makelov, A., Schmidt, L., Tsipras, D., and Vladu, A.
\newblock Towards deep learning models resistant to adversarial attacks.
\newblock \emph{arXiv preprint arXiv:1706.06083}, 2017.

\bibitem[Nagrani et~al.(2021)Nagrani, Yang, Arnab, Jansen, Schmid, and
  Sun]{nagrani2021attention}
Nagrani, A., Yang, S., Arnab, A., Jansen, A., Schmid, C., and Sun, C.
\newblock Attention bottlenecks for multimodal fusion.
\newblock \emph{Advances in Neural Information Processing Systems},
  34:\penalty0 14200--14213, 2021.

\bibitem[Oord et~al.(2018)Oord, Li, and Vinyals]{oord2018representation}
Oord, A. v.~d., Li, Y., and Vinyals, O.
\newblock Representation learning with contrastive predictive coding.
\newblock \emph{arXiv preprint arXiv:1807.03748}, 2018.

\bibitem[Park et~al.(2019)Park, Chan, Zhang, Chiu, Zoph, Cubuk, and
  Le]{park2018specaug}
Park, D.~S., Chan, W., Zhang, Y., Chiu, C.-C., Zoph, B., Cubuk, E.~D., and Le,
  Q.~V.
\newblock {SpecAugment}: A simple data augmentation method for automatic speech
  recognition.
\newblock In \emph{Interspeech 2019}. {ISCA}, sep 2019.
\newblock \doi{10.21437/interspeech.2019-2680}.
\newblock URL \url{https://doi.org/10.21437%2Finterspeech.2019-2680}.

\bibitem[Pham et~al.(2021)Pham, Dai, Ghiasi, Kawaguchi, Liu, Yu, Yu, Chen,
  Luong, Wu, Tan, and Le]{Pham2021}
Pham, H., Dai, Z., Ghiasi, G., Kawaguchi, K., Liu, H., Yu, A.~W., Yu, J., Chen,
  Y.-T., Luong, M.-T., Wu, Y., Tan, M., and Le, Q.~V.
\newblock Combined scaling for open-vocabulary image classification, 2021.
\newblock URL \url{https://arxiv.org/abs/2111.10050}.

\bibitem[Radford et~al.(2019)Radford, Wu, Child, Luan, Amodei, Sutskever,
  et~al.]{radford2019language}
Radford, A., Wu, J., Child, R., Luan, D., Amodei, D., Sutskever, I., et~al.
\newblock Language models are unsupervised multitask learners.
\newblock \emph{OpenAI blog}, 1\penalty0 (8):\penalty0 9, 2019.

\bibitem[Radford et~al.(2021)Radford, Kim, Hallacy, Ramesh, Goh, Agarwal,
  Sastry, Askell, Mishkin, Clark, Krueger, and Sutskever]{radford2021clip}
Radford, A., Kim, J.~W., Hallacy, C., Ramesh, A., Goh, G., Agarwal, S., Sastry,
  G., Askell, A., Mishkin, P., Clark, J., Krueger, G., and Sutskever, I.
\newblock Learning transferable visual models from natural language
  supervision.
\newblock \emph{ICML}, 2021.

\bibitem[Rebuffi et~al.(2017)Rebuffi, Kolesnikov, Sperl, and
  Lampert]{rebuffi2017icarl}
Rebuffi, S.-A., Kolesnikov, A., Sperl, G., and Lampert, C.~H.
\newblock icarl: Incremental classifier and representation learning.
\newblock In \emph{Proceedings of the IEEE conference on Computer Vision and
  Pattern Recognition}, pp.\  2001--2010, 2017.

\bibitem[Recht et~al.(2019)Recht, Roelofs, Schmidt, and
  Shankar]{recht2019imagenet}
Recht, B., Roelofs, R., Schmidt, L., and Shankar, V.
\newblock Do imagenet classifiers generalize to imagenet?
\newblock In \emph{ICML}, pp.\  5389--5400. PMLR, 2019.

\bibitem[Shankar et~al.(2021)Shankar, Dave, Roelofs, Ramanan, Recht, and
  Schmidt]{Shankar_2021_ICCV}
Shankar, V., Dave, A., Roelofs, R., Ramanan, D., Recht, B., and Schmidt, L.
\newblock Do image classifiers generalize across time?
\newblock In \emph{Proceedings of the IEEE/CVF International Conference on
  Computer Vision (ICCV)}, pp.\  9661--9669, October 2021.

\bibitem[Singh et~al.(2021)Singh, Hu, Goswami, Couairon, Galuba, Rohrbach, and
  Kiela]{singh2021flava}
Singh, A., Hu, R., Goswami, V., Couairon, G., Galuba, W., Rohrbach, M., and
  Kiela, D.
\newblock Flava: A foundational language and vision alignment model, 2021.
\newblock URL \url{https://arxiv.org/abs/2112.04482}.

\bibitem[Sohn(2016)]{sohn2016npair}
Sohn, K.
\newblock Improved deep metric learning with multi-class n-pair loss objective.
\newblock In Lee, D., Sugiyama, M., Luxburg, U., Guyon, I., and Garnett, R.
  (eds.), \emph{Advances in Neural Information Processing Systems}, volume~29.
  Curran Associates, Inc., 2016.
\newblock URL
  \url{https://proceedings.neurips.cc/paper/2016/file/6b180037abbebea991d8b1232f8a8ca9-Paper.pdf}.

\bibitem[Szegedy et~al.(2013)Szegedy, Zaremba, Sutskever, Bruna, Erhan,
  Goodfellow, and Fergus]{szegedy2013intriguing}
Szegedy, C., Zaremba, W., Sutskever, I., Bruna, J., Erhan, D., Goodfellow, I.,
  and Fergus, R.
\newblock Intriguing properties of neural networks.
\newblock \emph{arXiv preprint arXiv:1312.6199}, 2013.

\bibitem[Taori et~al.(2020)Taori, Dave, Shankar, Carlini, Recht, and
  Schmidt]{taori2020}
Taori, R., Dave, A., Shankar, V., Carlini, N., Recht, B., and Schmidt, L.
\newblock Measuring robustness to natural distribution shifts in image
  classification.
\newblock In Larochelle, H., Ranzato, M., Hadsell, R., Balcan, M., and Lin, H.
  (eds.), \emph{Advances in Neural Information Processing Systems}, volume~33,
  pp.\  18583--18599. Curran Associates, Inc., 2020.
\newblock URL
  \url{https://proceedings.neurips.cc/paper/2020/file/d8330f857a17c53d217014ee776bfd50-Paper.pdf}.

\bibitem[Tong et~al.(2022)Tong, Song, Wang, and Wang]{tong2022videomae}
Tong, Z., Song, Y., Wang, J., and Wang, L.
\newblock Videomae: Masked autoencoders are data-efficient learners for
  self-supervised video pre-training.
\newblock \emph{arXiv preprint arXiv:2203.12602}, 2022.

\bibitem[Vaswani et~al.(2017)Vaswani, Shazeer, Parmar, Uszkoreit, Jones, Gomez,
  Kaiser, and Polosukhin]{vaswani2017attention}
Vaswani, A., Shazeer, N., Parmar, N., Uszkoreit, J., Jones, L., Gomez, A.~N.,
  Kaiser, {\L}., and Polosukhin, I.
\newblock Attention is all you need.
\newblock \emph{NeurIPS}, 30, 2017.

\bibitem[Wang et~al.(2022)Wang, Bao, Dong, Bjorck, Peng, Liu, Aggarwal,
  Mohammed, Singhal, Som, et~al.]{wang2022image}
Wang, W., Bao, H., Dong, L., Bjorck, J., Peng, Z., Liu, Q., Aggarwal, K.,
  Mohammed, O.~K., Singhal, S., Som, S., et~al.
\newblock Image as a foreign language: Beit pretraining for all vision and
  vision-language tasks.
\newblock \emph{arXiv preprint arXiv:2208.10442}, 2022.

\bibitem[Wortsman et~al.(2022)Wortsman, Ilharco, Kim, Li, Kornblith, Roelofs,
  Lopes, Hajishirzi, Farhadi, Namkoong, and Schmidt]{Wortsman_2022_CVPR}
Wortsman, M., Ilharco, G., Kim, J.~W., Li, M., Kornblith, S., Roelofs, R.,
  Lopes, R.~G., Hajishirzi, H., Farhadi, A., Namkoong, H., and Schmidt, L.
\newblock Robust fine-tuning of zero-shot models.
\newblock In \emph{Proceedings of the IEEE/CVF Conference on Computer Vision
  and Pattern Recognition (CVPR)}, pp.\  7959--7971, June 2022.

\bibitem[Wu et~al.(2022{\natexlab{a}})Wu, Jastrzebski, Cho, and
  Geras]{wu2022characterizing}
Wu, N., Jastrzebski, S., Cho, K., and Geras, K.~J.
\newblock Characterizing and overcoming the greedy nature of learning in
  multi-modal deep neural networks.
\newblock In \emph{International Conference on Machine Learning}, pp.\
  24043--24055. PMLR, 2022{\natexlab{a}}.

\bibitem[Wu et~al.(2022{\natexlab{b}})Wu, Chen, Zhang, Hui, Berg-Kirkpatrick,
  and Dubnov]{wu2022large}
Wu, Y., Chen, K., Zhang, T., Hui, Y., Berg-Kirkpatrick, T., and Dubnov, S.
\newblock Large-scale contrastive language-audio pretraining with feature
  fusion and keyword-to-caption augmentation.
\newblock \emph{arXiv preprint arXiv:2211.06687}, 2022{\natexlab{b}}.

\bibitem[Xu et~al.(2022)Xu, Li, Baevski, Auli, Galuba, Metze, Feichtenhofer,
  et~al.]{xu2022masked}
Xu, H., Li, J., Baevski, A., Auli, M., Galuba, W., Metze, F., Feichtenhofer,
  C., et~al.
\newblock Masked autoencoders that listen.
\newblock \emph{arXiv preprint arXiv:2207.06405}, 2022.

\bibitem[Yang et~al.(2022)Yang, Fang, Zhu, Pryzant, Chen, Shi, Xu, Qian, Gao,
  Chen, et~al.]{yang2022code}
Yang, Z., Fang, Y., Zhu, C., Pryzant, R., Chen, D., Shi, Y., Xu, Y., Qian, Y.,
  Gao, M., Chen, Y.-L., et~al.
\newblock i-code: An integrative and composable multimodal learning framework.
\newblock \emph{arXiv preprint arXiv:2205.01818}, 2022.

\bibitem[You et~al.(2022)You, Zhou, Xiao, Codella, Cheng, Xu, Chang, and
  Yuan]{you2022msclip}
You, H., Zhou, L., Xiao, B., Codella, N., Cheng, Y., Xu, R., Chang, S.-F., and
  Yuan, L.
\newblock Learning visual representation from modality-shared contrastive
  language-image pre-training, 2022.
\newblock URL \url{https://arxiv.org/abs/2207.12661}.

\bibitem[Yuan et~al.(2021)Yuan, Chen, Chen, Codella, Dai, Gao, Hu, Huang, Li,
  Li, et~al.]{yuan2021florence}
Yuan, L., Chen, D., Chen, Y.-L., Codella, N., Dai, X., Gao, J., Hu, H., Huang,
  X., Li, B., Li, C., et~al.
\newblock Florence: A new foundation model for computer vision.
\newblock \emph{arXiv preprint arXiv:2111.11432}, 2021.

\bibitem[Zhang et~al.(2017)Zhang, Cisse, Dauphin, and
  Lopez-Paz]{zhang2017mixup}
Zhang, H., Cisse, M., Dauphin, Y.~N., and Lopez-Paz, D.
\newblock mixup: Beyond empirical risk minimization, 2017.
\newblock URL \url{https://arxiv.org/abs/1710.09412}.

\end{thebibliography}
\bibliographystyle{icml2022}

\clearpage

\appendix
{\Large {\bf Appendix}}\label{sec:appendix}

\section{Multimodal Self-supervised Models}\label{sec:ssl_methods}

\begin{figure}[t]
    \centering
    \includegraphics[width=0.8\columnwidth]{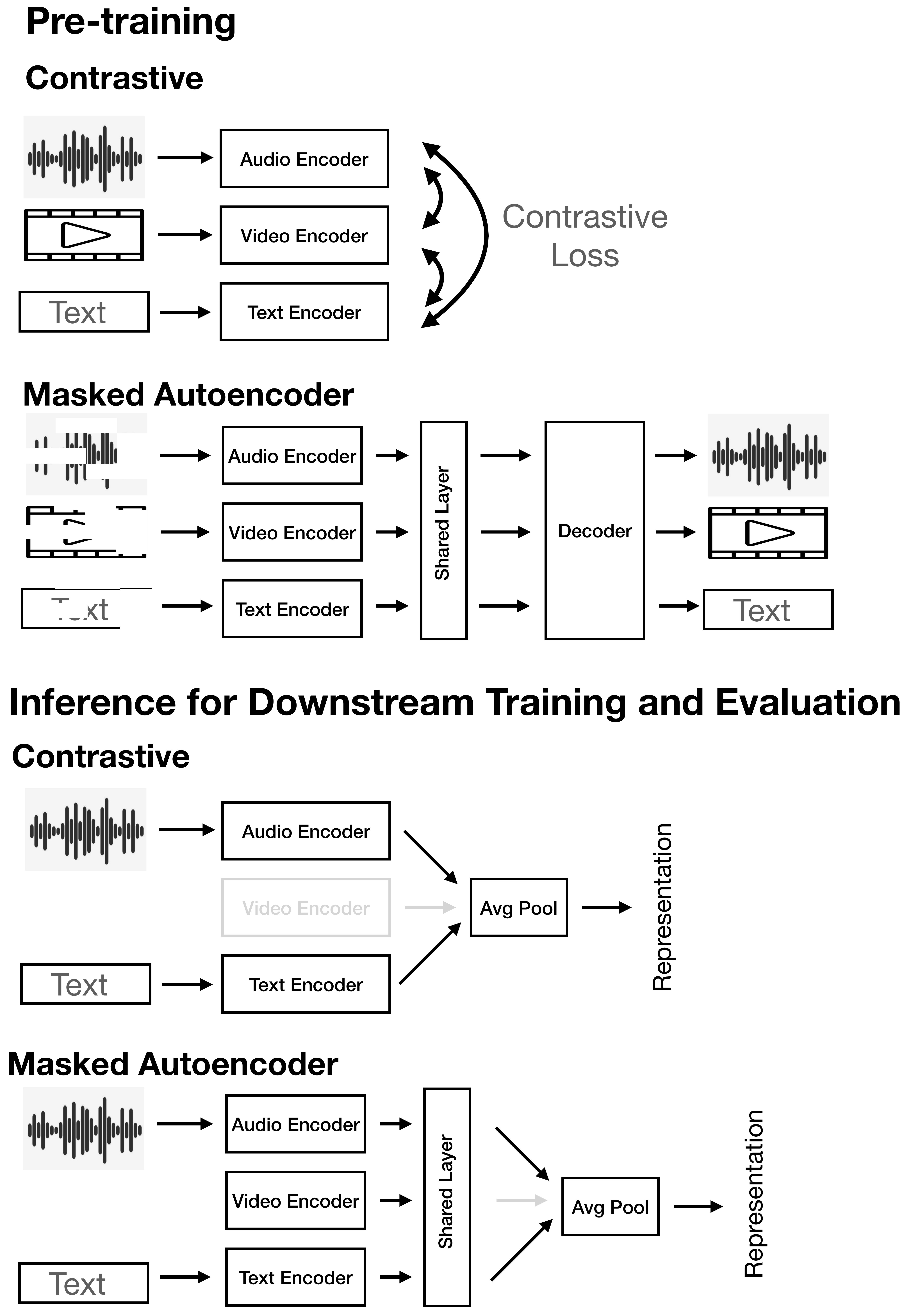}
    \caption{{\bf Diagram of pretraining and inference setup.} Top: We investigate two SSL losses during pretraining, either contrastive losses or masked autoencoders, to learn a multimodal representations. Bottom: We use this representation for a downstream task. Note that we can use this model to feed a subset of the pre-training modalities, in this figure using audio and text as an illustrative example.}
    \label{fig:ssl_methods}
\end{figure}

Contrastive learning for multiple modalities has been applied primarily for image and text~\cite{radford2021clip, jia2021scaling} using the NCE loss~\cite{gutmann2010noise,oord2018representation} per batch to embed paired examples from different modalities close to each other in a common latent Euclidean space. We follow this setup closely where we use $n$ modality specific encoders (in our case $n=3$ with text, video, and audio) and formulate ${n\choose 2}$ contrastive losses\footnote{We use the same global batch-contrastive loss as defined in \cite{radford2021clip}.} (see Fig.~\ref{fig:ssl_methods}, top).

Masked autoencoders (MAE), however, have been applied to setups beyond image and text, such as video~\cite{geng2022multimodal, georgescu2022audiovisual,girdhar2022omnimae,tong2022videomae} and audio~\cite{georgescu2022audiovisual,xu2022masked}. These methods learn to embed (masked) inputs in a latent space from which the original unmasked inputs can be reconstructed. In the case of multimodal inputs, the model learns to reconstruct each modality from a masked version of all modalities, and thus ideally encourage cross-modal interactions (see Fig.~\ref{fig:ssl_methods}, middle).

\paragraph{Downstream Training and Inference} For a downstream task we use the self-supervised learned model to compute a representation of multimodal inputs; this representation in turn is used for the task. We would like these models to produce a representation in the same latent space independent of whether they get as an input a single or many modalities. 

We can easily achieve this by applying average pooling across modality specific encoders (see Fig.~\ref{fig:ssl_methods}, bottom). In particular, denote by $E_i$ the encoder for modality $m_i$ that embeds an input $x|_{m_i}$ of this modality into a Euclidean space $E_i(x|_{m_i})\in\mathbb{R}^d$ (see Sec.~\ref{sec:notation} for notation). Suppose, at downstream training or inference time, the data $D|_{M'}$ have a subset of modalities $M'\subseteq M$. Then, the final representation of these data is for $x\in D$:
\begin{equation}
    E(x) = \frac{1}{|M'|}\sum_{m'\in M'} E_{m'}(x|_{m'})
\end{equation}

\section{AudioSet Details}\label{subsec:audioset-details}

The number of segments in the AudioSet downloads for unbalanced train, balanced train, and evaluation are 2,042,985 examples, 22,176 examples, and 20,383 examples, respectively \footnote{\url{https://research.google.com/audioset}}. Since YouTube videos can be removed over time, it is common that not all examples can be downloaded from the provided URLs in the dataset. For the unbalanced train, balanced train, and evaluation, we were able to obtain 1,743,790 examples (86.7\%), 18,649 (84.1\%), and 17,065 (83.7\%) examples, respectively.

One concern we had with using the video title as an input modality was whether the titles simply contain the label. If this were the case, the model could trivially solve the task by just looking at the text. It is true that the authors of AudioSet noted that the videos selected for human annotation were guided by an internal video-level automatic annotation system and a metadata-based approach that included the video titles. This means the labels and title for a given example are undoubtedly correlated, but cursory inspection of the examples reveals that the titles are still a rather noisy source of information with respect to the classification task. For example, see some randomly drawn samples from the evaluation set in ~\ref{tab:audioset-examples}. Furthermore, metadata like titles are abundant in webcrawled data and obtaining them is a substantially cheaper process than obtaining human annotations. For these reasons, we believed that utilizing this textual information was reasonable/justified for the robustness analyses presented in this paper.

\begin{table*}[ht!]
    \centering
    {\scriptsize
    \begin{tabular}{|l | l |}
    \hline 
    Text & Labels \\ \hline 
        \hline 
    Lps: More Than That [7] (Season 3 Finale Part 1)  {Christmas Special}
        & Brass instrument Clarinet \\
        \hline 
    Muzik tipiko di Korsou/ Traditional Curacao music 
        & Flamenco, Music, Mandolin, Music of Latin America \\
        \hline 
    Tie Down Roping - 2013 NFR Round 8
        & Bang \\
        \hline 
    COUPLES YOGA CHALLENGE 
        & Music, Speech, Breathing \\
        \hline 
    Eventide Timefactor Delay Pedal Part 2 
        & Effects unit, Guitar, Music, Musical instrument, Chorus effect, Plucked string instrument \\
        \hline 
    Fill Your Bucket - Children's Song by The Learning Station 
        & Jingle (music), Music \\
        \hline 
    Klakson, który zwala z nóg
        & Vehicle, ``Vehicle horn, car horn, honking'', Speech, ``Outside, urban or manmade'' \\
        \hline 
    A Cappella Pitch Perfect Mashup 
        & Singing, Music, Choir, Vocal music, A capella \\
        \hline 
    John Lennon - Imagine Goat Edition 
        & Music, Independent music, Song, Sheep, Bleat \\
        \hline 
    Weird Or What? - The Bloop - World Mysteries 
        & Music, Rumble, Speech \\
    \hline 
    \end{tabular}
    \caption{Random examples of text and associated labels from AudioSet evaluation set.\label{tab:audioset-examples}}}
\end{table*}

\section{Training Hyperparameters}\label{sec:hyperparameters}

 \textbf{Pretraining}. For \textit{audio}, \textit{video}, and \textit{text}, we learn a representation using Contrastive Learning and MAE on the unbalanced training set of AudioSet with a global batch of 1024, the AdamW optimizer~\cite{loshchilov2017decoupled}, and a learning rate of 8e-4. We train the contrastive model for 32 epochs (54K steps) and MAE for 256 epochs (435K steps). We run the downstream training 30 additional epochs. Note that before we train the full model we learn a linear classifier on top of the frozen pretrained weights (referred to as linear probing). The trained classifier weights are then used for initializing the classifier at the beginning of full downstream training, which we found to be crucial for achieving good finetuning performance. Following \cite{he2022masked}, during linear probing we include BatchNorm without the affine transformation before the final classifier.

\textbf{Linear Probing}. When linear probing, we first precompute the frozen backbone's features and reuse those for subsequent epochs. When precomputing features, we do not use any of the random data augmentations (the data augmentations are the same fixed augmentations applied during evaluation).

\textbf{Finetuning}. When finetuning models with the distillation loss (MASD), we use a loss weight of 0.5 (equal weight). Although we also ablated choices regarding temperature, exponential moving average on the teacher, and randomly sampling student modalities each batch, none of these significantly improved results. Therefore, we don't use them in our final reported results and instead opt for the simplest setup. For more details of training hyperparameters, see ~\ref{tab:model-hyperparams}.



\begin{table}[ht!]
    \centering
    \begin{tabular}{|l |l|}
        \hline
        Model & Num Params \\
        \hline 
        \hline
        AV-MAE & 191M \\ 
        \hline 
        AV-Contrastive & 173M \\ 
        \hline 
        AV-Contrastive (CLIP Init.) & 172M \\ \hline
        \hline 
        AVT-MAE & 334M \\ 
        AVT-Contrastive & 297M \\ 
        AVT-Contrastive (CLIP Init.) & 236M \\ 
        \hline
    \end{tabular}
    
    \caption{Number of parameters in each model configuration. The reason AVT-Contrastive with CLIP initialization has fewer parameters than AVT-Contrastive is that we use the same transformer architecture for all modalities by default, whereas CLIP \cite{radford2021clip} has a smaller text encoder than vision encoder.}
    \label{tab:parameter-counts}
\end{table}

\begin{table}[ht!]
    \centering
    \small
    \begin{tabular}{|l|l|}
        \hline
        Name & Value \\ \hline 
        \hline 
        Video sampling stride & 32 \\
        \hline 
        Video max sampled frames & 8 \\ 
        \hline 
        Video spatial size & 224 \\
        \hline 
        Video mean & 0.45 \\
        \hline 
        Video std & 0.225 \\
        \hline 
        \hline
        Audio seconds sampled & 8 \\
        \hline 
        Audio mel bins & 128 \\ 
        \hline 
        Audio mean &  -4.2677393 \\
        \hline 
        Audio std & 4.5689974 \\
        \hline 
        Audio original sample frequency & 44.1 kHz \\
        \hline 
        Audio resampled frequency & 16 kHz \\
        \hline 
        \hline 
        Text max sequence length (BPE tokens) & 60 \\
        \hline 
        Text vocab size & 50262 \\
        \hline
    \end{tabular}
    \caption{Main data hyperparameters for each modality.}
    \label{tab:data-hyperparams}
\end{table}

\begin{table}[h]
    \centering\tiny
    \begin{tabular}{|c |c|c |c|c |c|c |}
        \hline
        Config & \multicolumn{2}{c|}{Pretraining} & \multicolumn{2}{c|}{Linear Probing} & \multicolumn{2}{c|}{Finetuning}\\ 
\cline{2-7}
        & Contr. & MAE  & Contr. & MAE & Contr. & MAE \\ \hline
        \hline 
        global batch 
            & 1024 & 1024 & 256 & 128 & 128 & 64 \\
        \hline 
        learning rate 
            & 8e-4 & 8e-4 & 1e-2 & 1e-2 & 1e-4 & 1e-4 \\
        \hline 
        LR warmup
            & 1000 & 2000 & 200 & 200 & 1000 & 2000 \\
        \hline 
        epochs 
            & 32 & 256 & 360 & 360 & 30 & 60 \\
        \hline 
        optimizer 
            & AdamW & AdamW & AdamW & AdamW & AdamW & AdamW \\
        \hline
    \end{tabular}
    
    \caption{Main training hyperparameters used for pretraining, linear probing, and finetuning, for both contrastive and MAE.}
    \label{tab:model-hyperparams}
\end{table}

\section{Architecture Details}\label{subsec:architecture-details}

\textbf{Contrastive}. Our contrastive model is initialized with pretrained CLIP ViT-B/16 weights \cite{radford2021clip}. The original CLIP model consists of two separate encoders and is intended for images and text, while our model has three encoders and is intended for audio, videos, and text. We use the same model code provided in the official CLIP GitHub repository\footnote{\url{https://github.com/openai/CLIP}}. For text, we load the CLIP text encoder as-is. For video and audio, we need to make small modifications to the positional encodings to account for the differences compared to images. For video, we adopt the separable positional encoding as described in \cite{feichtenhofer2022masked} and initialize the spatial component with the weights from CLIP's image encoder. For audio, we perform bilinear interpolation of the positional encodings \cite{dosovitskiy2020image} in order to accommodate the input audio shape of $800 \times 128$.

\textbf{MAE}. Due to architectural differences, we cannot easily initialize the MAE models from CLIP, so those models are pretrained from scratch\footnote{This is partially why MAE is pretrained for 256 epochs, whereas the contrastive models are pretrained for 32 epochs. Another reason our contrastive method is pretrained for fewer epochs is because it was challenging to avoid overfitting if we pretrained any longer.}. Following \cite{gong2022contrastive}, our MAE consists of modality-separate encoders with the ViT-B/16 architectures, but where the final (12th) layer is shared across modalities. We ran ablations for the number of shared layers and found that one modality-shared layer yields the best results, similar to \cite{gong2022contrastive}. In fact, downstream training on top of a MAE trained model \textit{decreases} performance and robustness as we increase the number of modality-shared encoder layers. For audio, we use a fixed 2D sinusoidal positional encoding as described in \cite{huang2022audiomae} and \cite{he2022masked}. For text, we use fixed 1D sinusoidal positional encodings as described in \cite{vaswani2017attention}. 

\vspace{-0.3cm}
\section{Data Augmentations}
The main data hyperparameters for each modality are outlined in table ~\ref{tab:data-hyperparams}. Overall, we aim to largely reuse established data pipelines for each modality, following \cite{huang2022audiomae} for audio, \cite{feichtenhofer2022masked} for video, and \cite{radford2019language} for text. This also includes applying mixup \cite{zhang2017mixup} with rate $0.5$ on all inputs/labels except text, drop path \cite{huang2016droppath} with drop rate $0.1$., and SpecAug \cite{park2018specaug} with time/frequency masking of 192/48. 

For videos, during both pretraining and downstream training we also use color augmentations for brightness (max delta = 0.2), contrast (max delta=0.1), saturation (max delta=0.0), hue (max delta=0.025). Also, we ensure the 8 seconds of audio/video are aligned such that the audio segment begins at the first sampled video frame and ends at the last sampled video frame.

For pretraining MAE, we follow \cite{feichtenhofer2022masked} and adopt repeated sampling, where each batch is duplicated/repeated some number of times (for us, we set the number of repeats per batch to 2), which improves training throughput due to the high cost of loading audio/video. This only makes sense for MAE with high masking ratios (we mask out 80\% of the audio and 90\% of the video during pretraining).

For ImageNet-Captions we apply standard RGB normalization and take a center-crop. No other augmentations are used.

\vspace{-0.2cm}
\section{Complete Experiments}
\vspace{-0.2cm}
\label{sec:complete-results}
For each SSL method, we pretrain one backbone model on AudioSet-2M using all three available input modalities. We then run linear probing and finetuning separately on all unique combinations of modalities. For AudioSet there are seven total possible combinations of modalities: 
\[
\{A\}, \{V\}, \{T\}, \{AT\}, \{AV\}, \{VT\}, \{AVT\} 
\]
where $A=\textrm{audio}$, $V=\textrm{video}$, $T=\textrm{text}$.

For Kinetics-400 there are three possible combinations:
\[
\{A\}, \{V\}, \{AV\}
\]

For ImageNet-Captions there are three possible combinations as well:
\[
\{I\}, \{T\}, \{IT\}
\]
where $I=\textrm{image}$ and $T=\textrm{text}$.

Finally, we test each produced model on all possible input modality combinations. 

Since we have two representation learning techniques, \textit{Contrastive} and \textit{MAE}, and for each of them we are to perform \textit{linear probe downstream training}, \textit{full model downstream training}, and \textit{MASD}, this results in $2 \times 3 \times 7 \times 7 = 294$ different mAP values corresponding to all possible combinations of the above. While in the main paper we show various aggregates the complete results for AudioSet are in Table~\ref{tab:complete results}.

Similarly, for Kinetics-400 and ImageNet-Captions we have $2 \times 3 \times 3 \times 3 = 54$ experiments each listed in Table~\ref{tab:complete_results_kinetics} and Table~\ref{tab:complete_results_incaptions}.

\vspace{-0.3cm}
\section{Training/Evaluation Combinations}
\vspace{-0.2cm}
In Sec.~\ref{subsec:metrics} we introduce metrics over training $M_T$ and evaluation $M_E$ modality sets over several types of combinations. For the sake of clarity, we list them in Table~\ref{tab:training_evaluation_modality_combos} explicitly for the case of \textit{audio}, \textit{video}, and \textit{text}.

\begin{table}[h]
    \centering\tiny
    \begin{tabular}{|c|c|c|c|c|c|c|c|}
    \hline
    & \multicolumn{7}{c|}{Training Modality Set} \\
    \cline{2-8}
         & V & A & T & AV & AT & VT & AVT \\
        \hline
        \hline
        \multirow{2}{*}{\shortstack{Missing \\ at Test}} & & & & A, V& A, T & V, T & A, V, T, \\
         & & & & & & & AV, AT, VT \\
        \hline
        \multirow{2}{*}{\shortstack{Added \\ at Test}} & VA, VT, & AT, AV, & AT, VT, & AVT & AVT & AVT &  \\
       & AVT &  AVT &  AVT &  &  &  &  \\
        \hline
        \multirow{2}{*}{\shortstack{Transfer- \\ability}} & A, T, & V, T, & A, V, & T& V & A &  \\
         & AT & VT & AV & & & &  \\
        \hline
    \end{tabular}
    \caption{Combinations of training and evaluation modality sets. For each training modality set $M_T$, we list all possible evaluation modality set $M_E$.}
    \label{tab:training_evaluation_modality_combos}
\end{table}

\begin{table*}[t]
    \centering
    {\scriptsize
    \begin{tabular}{|c|c|c|c|c|c|c|c|c|c|c|c|c|c|c|}
\hline
Train mod. & \multicolumn{7}{c|}{V} & \multicolumn{7}{c|}{A}\\ 
\cline{1-15}
 Test mod. & V & A & T & AV & AT & VT & AVT & V & A & T & AV & AT & VT & AVT\\ 
\hline
\hline
Contr., MASD + WiseFT & 30.0 & 24.6 & 22.6 & 30.7 & 29.6 & 29.5 & 32.4 & 20.9 & 39.5 & 23.7 & 40.9 & 41.6 & 28.3 & 41.8\\ 
\hline
Contr., MASD & 25.5 & 24.3 & 21.9 & 29.9 & 28.7 & 29.0 & 31.6 & 20.9 & 39.5 & 23.3 & 41.1 & 41.7 & 28.1 & 42.0\\ 
\hline
Contr., WiseFT & 26.0 & 8.6 & 12.7 & 26.7 & 15.6 & 27.4 & 27.9 & 4.5 & 39.3 & 11.3 & 35.3 & 38.6 & 10.7 & 34.7\\ 
\hline
Contr., FT & 25.6 & 10.3 & 14.3 & 26.4 & 17.9 & 26.9 & 27.9 & 5.2 & 39.5 & 12.4 & 35.7 & 38.9 & 12.0 & 35.3\\ 
\hline
Contr., LP & 24.9 & 10.5 & 14.3 & 25.7 & 18.3 & 26.1 & 27.1 & 6.2 & 36.5 & 13.5 & 32.1 & 35.2 & 13.9 & 32.0\\ 
\hline
Contr., FT on 2M & 27.4 & 23.8 & 20.2 & 34.2 & 29.3 & 31.7 & 36.2 & 15.3 & 39.4 & 17.6 & 41.3 & 42.1 & 21.5 & 42.2\\ 
\hline
MAE, MASD & 18.1 & 17.9 & 12.2 & 21.8 & 24.0 & 21.3 & 24.9 & 10.5 & 34.8 & 15.4 & 36.3 & 38.2 & 21.0 & 38.5\\ 
\hline
MAE, FT & 17.7 & 0.8 & 1.9 & 16.3 & 1.6 & 18.7 & 17.2 & 0.7 & 34.4 & 1.4 & 33.7 & 35.2 & 1.3 & 34.6\\ 
\hline
MAE, LP & 12.7 & 0.7 & 1.5 & 1.9 & 1.4 & 5.9 & 3.2 & 0.8 & 31.4 & 2.5 & 16.0 & 12.6 & 3.3 & 11.4\\ 
\hline
\end{tabular}
\vspace{0.4cm} \\
    \begin{tabular}{|c|c|c|c|c|c|c|c|c|c|c|c|c|c|c|}
\hline
Train mod. & \multicolumn{7}{c|}{T} & \multicolumn{7}{c|}{AV}\\ 
\cline{1-15}
 Test mod. & V & A & T & AV & AT & VT & AVT & V & A & T & AV & AT & VT & AVT\\ 
\hline
\hline
Contr., MASD + WiseFT & 23.1 & 29.7 & 30.3 & 33.3 & 37.2 & 33.8 & 38.1 & 22.2 & 38.2 & 25.1 & 44.2 & 42.8 & 29.9 & 44.9\\ 
\hline
Contr., MASD & 23.02 & 29.7 & 30.6 & 33.2 & 37.0 & 34.1 & 38.2 & 20.9 & 38.0 & 24.4 & 44.2 & 42.8 & 29.0 & 45.4\\ 
\hline
Contr., WiseFT & 15.2 & 17.2 & 30.4 & 24.3 & 34.3 & 31.9 & 35.2 & 23.2 & 38.1 & 19.7 & 44.4 & 41.4 & 28.8 & 44.3\\ 
\hline
Contr., FT & 14.8 & 16.5 & 31.7 & 23.9 & 35.9 & 33.9 & 37.1 & 22.5 & 38.4 & 20.8 & 43.7 & 41.5 & 28.8 & 43.9\\ 
\hline
Contr., LP & 16.2 & 19.0 & 28.1 & 25.5 & 32.0 & 29.0 & 32.8 & 23.0 & 35.3 & 21.2 & 39.8 & 37.4 & 28.6 & 39.5\\ 
\hline
Contr., FT on 2M & 18.3 & 25.2 & 30.0 & 28.8 & 37.7 & 34.1 & 39.1 & 23.6 & 36.3 & 18.6 & 45.6 & 40.2 & 28.7 & 44.8\\ 
\hline
MAE, MASD & 9.0 & 21.5 & 25.2 & 25.8 & 32.2 & 28.2 & 33.5 & 8.9 & 31.7 & 18.2 & 39.5 & 36.2 & 20.9 & 40.9\\ 
\hline
MAE, FT & 0.6 & 0.9 & 24.7 & 0.8 & 22.5 & 24.1 & 22.1 & 9.1 & 31.9 & 1.4 & 39.8 & 33.1 & 11.3 & 40.1\\ 
\hline
MAE, LP & 0.7 & 1.0 & 22.8 & 0.9 & 14.5 & 16.9 & 13.3 & 6.8 & 29.0 & 3.2 & 34.7 & 13.0 & 8.3 & 18.6\\ 
\hline
\end{tabular}
\vspace{0.4cm} \\
    \begin{tabular}{|c|c|c|c|c|c|c|c|c|c|c|c|c|c|c|}
\hline
Train mod. & \multicolumn{7}{c|}{AT} & \multicolumn{7}{c|}{VT}\\ 
\cline{1-15}
 Test mod. & V & A & T & AV & AT & VT & AVT & V & A & T & AV & AT & VT & AVT\\ 
\hline
\hline
Contr., MASD + WiseFT & 23.8 & 38.6 & 27.3 & 42.4 & 47.7 & 33.6 & 47.9 & 24.8 & 30.2 & 28.7 & 32.9 & 37.3 & 36.4 & 39.8\\ 
\hline
Contr., MASD & 24.1 & 38.3 & 26.1 & 42.4 & 48.3 & 33.7 & 48.8 & 24.0 & 30.3 & 28.4 & 32.2 & 37.3 & 36.5 & 40.0\\ 
\hline
Contr., WiseFT & 10.7 & 38.5 & 28.4 & 39.7 & 47.5 & 28.1 & 45.8 & 24.1 & 13.4 & 27.9 & 29.1 & 30.6 & 36.6 & 38.2\\ 
\hline
Contr., FT & 11.0 & 36.5 & 29.4 & 38.1 & 47.7 & 30.0 & 46.4 & 23.7 & 16.1 & 30.1 & 30.0 & 33.2 & 36.7 & 38.7\\ 
\hline
Contr., LP & 12.8 & 35.6 & 26.4 & 36.2 & 41.2 & 26.1 & 39.6 & 24.5 & 16.3 & 27.0 & 30.2 & 29.7 & 33.0 & 35.2\\ 
\hline
Contr., FT on 2M & 13.8 & 35.4 & 29.4 & 38.7 & 49.3 & 32.4 & 47.7 & 21.8 & 22.9 & 28.2 & 31.6 & 36.9 & 37.4 & 42.3\\ 
\hline
MAE, MASD & 15.1 & 28.7 & 21.6 & 32.6 & 42.7 & 30.0 & 43.8 & 7.1 & 25.4 & 23.0 & 26.6 & 34.3 & 29.7 & 36.0\\ 
\hline
MAE, FT & 0.6 & 27.2 & 17.7 & 27.7 & 41.8 & 18.2 & 40.3 & 9.6 & 1.1 & 23.5 & 10.6 & 23.3 & 30.7 & 29.3\\ 
\hline
MAE, LP & 0.8 & 23.4 & 12.9 & 17.2 & 38.0 & 13.2 & 35.0 & 5.6 & 1.0 & 18.3 & 2.2 & 16.7 & 27.7 & 21.6\\ 
\hline
\end{tabular}
\vspace{0.4cm} \\
\begin{tabular}{|c|c|c|c|c|c|c|c|}
\hline
Train mod. & \multicolumn{7}{c|}{AVT}\\ 
\cline{1-8}
 Test mod. & V & A & T & AV & AT & VT & AVT\\ 
\hline
\hline
Contr., MASD + WiseFT & 22.3 & 37.2 & 27.5 & 43.4 & 47.1 & 34.7 & 49.1\\ 
\hline
Contr., MASD & 18.7 & 34.9 & 29.0 & 41.9 & 47.0 & 34.5 & 49.4\\ 
\hline
Contr., WiseFT & 22.3 & 37.2 & 27.5 & 43.4 & 47.1 & 34.7 & 49.1\\ 
\hline
Contr., FT & 18.7 & 34.9 & 29.0 & 41.9 & 47.0 & 34.5 & 49.4\\ 
\hline
Contr., LP & 22.1 & 34.5 & 26.5 & 39.2 & 40.3 & 31.6 & 41.5\\ 
\hline
Contr., FT on 2M & 19.3 & 34.6 & 27.5 & 42.5 & 47.7 & 35.5 & 51.9\\ 
\hline
MAE, MASD & 3.8 & 31.5 & 19.7 & 34.2 & 44.1 & 22.9 & 46.0\\ 
\hline
MAE, FT & 3.8 & 31.5 & 19.7 & 34.2 & 44.1 & 22.9 & 46.0\\ 
\hline
MAE, LP & 5.5 & 23.0 & 12.2 & 27.0 & 36.5 & 20.6 & 39.2\\ 
\hline
\end{tabular}
    
    \caption{Complete results on AudioSet used in the calculation of our metrics.}
    \label{tab:complete results}
    }
\end{table*}

\begin{table*}
    \centering
    
\begin{tabular}{|c|c|c|c|c|c|c|c|c|c|}
\hline
Train mod. & \multicolumn{3}{c|}{V} & \multicolumn{3}{c|}{A} & \multicolumn{3}{c|}{AV}\\ 
\cline{1-10}
 Test mod. & V & A & AV & V & A & AV & V & A & AV\\ 
\hline
\hline
Contr., MASD & 66.9 & 27.4 & 68.7 & 58.4 & 27.4 & 55.1 & 47.0 & 11.1 & 70.5\\ 
\hline
Contr., FT & 67.1 & 2.3 & 67.0 & 4.9 & 27.5 & 28.6 & 47.0 & 11.1 & 70.5\\ 
\hline
Contr., LP & 55.8 & 10.7 & 41.1 & 21.7 & 24.3 & 32.5 & 50.2 & 18.5 & 57.9\\ 
\hline
\end{tabular}

    \caption{Complete results used on Kinetics-400 in the calculation of our metrics.}
    \label{tab:complete_results_kinetics}
\end{table*}

\begin{table*}
    \centering

\begin{tabular}{|c|c|c|c|c|c|c|c|c|c|}
\hline
Train mod. & \multicolumn{3}{|c|}{I} & \multicolumn{3}{c|}{T} & \multicolumn{3}{c|}{IT}\\ 
\cline{1-10}
 Test mod. & I & T & IT & I & T & IT & I & T & IT\\ 
\hline
\hline
Contr., MASD & 84.72 & 82.93 & 92.42 & 84.4 & 82.94 & 93.17 & 70.54 & 79.38 & 93.9\\ 
\hline
Contr., FT & 84.61 & 66.1 & 89.06 & 54.82 & 82.91 & 87.14 & 70.54 & 79.38 & 93.9\\ 
\hline
Contr., LP & 82.35 & 45.57 & 72.29 & 43.6 & 79.11 & 83.68 & 72.19 & 75.06 & 92.01\\ 
\hline
\end{tabular}
    
    \caption{Complete results used on ImageNet-Captions in the calculation of our metrics.}
    \label{tab:complete_results_incaptions}
    
\end{table*}




\end{document}